\documentclass[lettersize,journal]{IEEEtran}

\usepackage{amsmath,amsfonts}
\usepackage{algorithmic}
\usepackage{algorithm}
\usepackage{array}
\usepackage[caption=false,font=normalsize,labelfont=sf,textfont=sf]{subfig}
\usepackage{textcomp}
\usepackage{stfloats}
\usepackage{url}
\usepackage{verbatim}
\usepackage{graphicx}
\usepackage{cite}
\usepackage{hyperref}
\hypersetup{
hidelinks,
colorlinks=true,
linkcolor=black,
citecolor=black
}
\usepackage{cuted}
\usepackage{capt-of}
\usepackage{colortbl}
\usepackage{xcolor}
\usepackage{array}
\usepackage{graphicx}
\usepackage{subcaption}
\usepackage{lipsum}
\usepackage{arydshln}
\usepackage{xspace}
\usepackage{multirow}
\usepackage{multicol}
\usepackage{booktabs} 
\usepackage{tabu}
\usepackage{makecell}
\usepackage{mwe}
\usepackage{bm}

\newcommand{\best}{\cellcolor{red!50}}
\newcommand{\sbest}{\cellcolor{orange!80}}
\newcommand{\tbest}{\cellcolor{yellow!80}}

\definecolor{r}{rgb}{0,0,0}
\usepackage{fancyhdr}

\fancypagestyle{firstpage}{
  \fancyhf{} 
  \fancyfoot[C]{\footnotesize 1077-2626 © 2024 IEEE. Personal use is permitted, but republication/redistribution requires IEEE permission. \\See https://www.ieee.org/publications/rights/index.html for more information.} 

}

\begin{document}

\title{PGSR: Planar-based Gaussian Splatting for Efficient and High-Fidelity Surface Reconstruction}

\author{
Danpeng Chen, \and Hai Li, \and Weicai Ye, \and Yifan Wang, \and Weijian Xie, \and Shangjin Zhai, \\
\and Nan Wang, \and Haomin Liu, \and Hujun Bao, \and Guofeng Zhang
\thanks{H. Bao, G. Zhang, W. Ye are with the State Key Lab of CAD\&CG, Zhejiang University. E-mails: \{baohujun, zhangguofeng\}@zju.edu.cn, maikeyeweicai@gmail.com.}
\thanks{D. Chen and W. Xie are with the State Key Lab of CAD\&CG, Zhejiang University. W. Xie is also affiliated with SenseTime Research, and D. Chen is also affiliated with Tetras.AI. E-mails: 11921155@zju.edu.cn, xieweijian@sensetime.com.}
\thanks{H. Li is with RayNeo. E-mail: lihai@ffalcon.cn.}
\thanks{Y. Wang is with Shanghai AI Laboratory. E-mail: wangyifan@pjlab.org.cn.}
\thanks{S. Zhai, N. Wang and H. Liu are with SenseTime Research. E-mails: \{zhaishangjin, wangnan, liuhaomin\}@sensetime.com.}
\thanks{Corresponding author: Guofeng Zhang}
\thanks{Digital Object Identifier 10.1109/TVCG.2024.3494046}
}

\maketitle
\thispagestyle{firstpage}
\begin{strip}\centering
\includegraphics[width=0.99\linewidth]{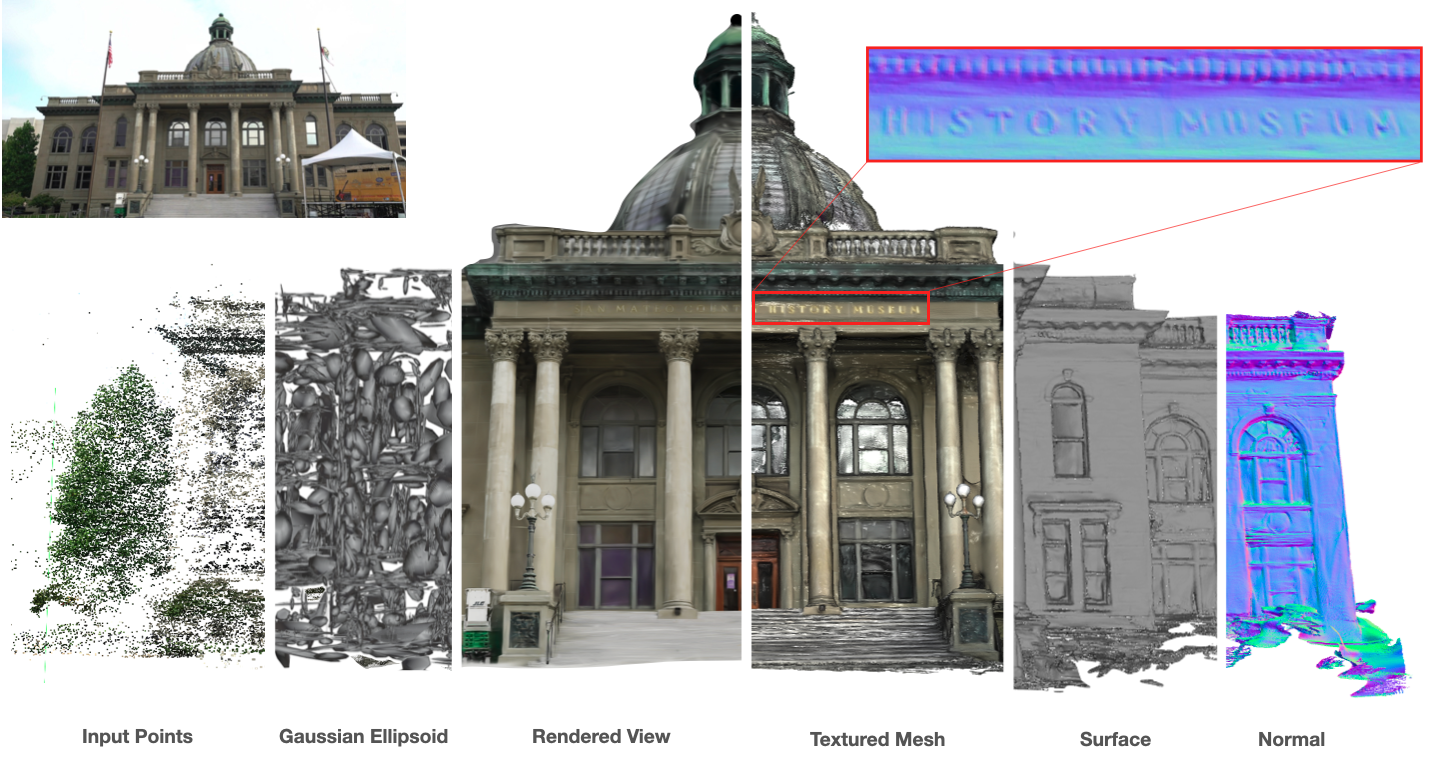}
\captionsetup{font={footnotesize}}
\captionof{figure}{
\textbf{PGSR representation.} We present a \textbf{P}lanar-based \textbf{G}aussian \textbf{S}platting \textbf{R}econstruction representation for efficient and high-fidelity surface reconstruction from multi-view RGB images without any geometric prior (depth or normal from pre-trained model). The courthouse reconstructed by our method demonstrates that PGSR can recover geometric details, such as textual details on the building. 
From left to right: input SfM points, planar-based Gaussian ellipsoid, rendered view, textured mesh, surface, and normal.
}
\label{fig:feature-graphic}
\end{strip}

\begin{abstract}
Recently, 3D Gaussian Splatting (3DGS) has attracted widespread attention due to its high-quality rendering, and ultra-fast training and rendering speed. However, due to the unstructured and irregular nature of Gaussian point clouds, it is difficult to guarantee geometric reconstruction accuracy and multi-view consistency simply by relying on image reconstruction loss. Although many studies on surface reconstruction based on 3DGS have emerged recently, the quality of their meshes is generally unsatisfactory. To address this problem, we propose a fast planar-based Gaussian splatting reconstruction representation (PGSR) to achieve high-fidelity surface reconstruction while ensuring high-quality rendering. Specifically, we first introduce an unbiased depth rendering method, which directly renders the distance from the camera origin to the Gaussian plane and the corresponding normal map based on the Gaussian distribution of the point cloud, and divides the two to obtain the unbiased depth. We then introduce single-view geometric, multi-view photometric, and geometric regularization to preserve global geometric accuracy. We also propose a camera exposure compensation model to cope with scenes with large illumination variations. 
Experiments on indoor and outdoor scenes show that the proposed method achieves fast training and rendering while maintaining high-fidelity rendering and geometric reconstruction, outperforming 3DGS-based and NeRF-based methods. Our code will be made publicly available, and more information can be found on our project page (\textcolor{magenta}{\href{https://zju3dv.github.io/pgsr/}{https://zju3dv.github.io/pgsr/}}).
\end{abstract}


\section{Introduction}

\begin{figure*}[htb]
    \centering
    \includegraphics[width=0.98\linewidth]{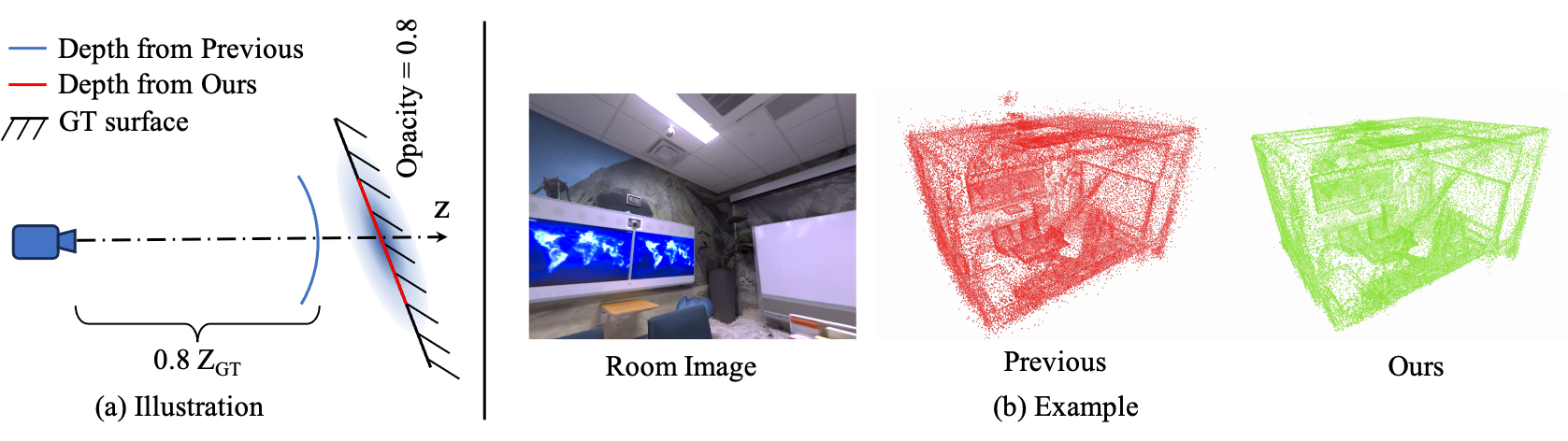}
    \captionsetup{font={footnotesize}}
    \caption{\textbf{Unbiased depth rendering.} (a) Illustration of the rendered depth: We take a single Gaussian, flatten it into a plane, and fit it onto the surface as an example. Our rendered depth is the intersection point of rays and surfaces, matching the actual surface. In contrast, the depth from previous methods~\cite{jiang2023gaussianshader, cheng2024gaussianpro} corresponds to a curved surface and may deviate from the actual surface. (b) We use true depth to supervise two different depth rendering methods. After optimization, we map the positions of all Gaussian points. Gaussians of our method fit well onto the actual surface, while the previous method results in noise and poor adherence to the surface.}
    \label{fig:unbias}
\end{figure*}

\IEEEPARstart{N}{ovel} view synthesis and geometry reconstruction are challenging and crucial tasks in computer vision, widely used in AR/VR~\cite{deng2022fov,ye2021superplane}, 3D content generation~\cite{tang2023dreamgaussian,poole2022dreamfusion, gao2024luminat2x, chen2024meshanything}, and autonomous driving. To achieve a realistic and immersive experience in AR/VR, novel view synthesis needs to be sufficiently convincing, and 3D reconstruction~\cite{li2020saliency,liu2021coxgraph, ye2023PVO, ye2022deflowslam} needs to be finely detailed. Recently, neural radiance fields\cite{mildenhall2021nerf, Ye2023IntrinsicNeRF, huang2024nerfdet++, ming2022idf} have been widely used to tackle this task, achieving high-fidelity novel view synthesis~\cite{muller2022instant,barron2021mip,barron2023zip} and 3D geometry reconstruction~\cite{wang2021neus,li2023neuralangelo}. However, due to the computationally intensive volume rendering methods, neural radiance fields often require training times of several hours to even hundreds of hours, and rendering speeds are difficult to achieve in real-time. Recently, 3D Gaussian Splatting (3DGS)~\cite{kerbl20233d} has made groundbreaking advancements in this field. By optimizing the positions, rotations, scales, and appearances of the explicit 3D Gaussians and combining alpha-blend rendering, 3DGS has achieved training times in the order of minutes and rendering speeds in the millisecond range.  

\begin{figure}[htb]
    \centering
    \includegraphics[width=1\linewidth]{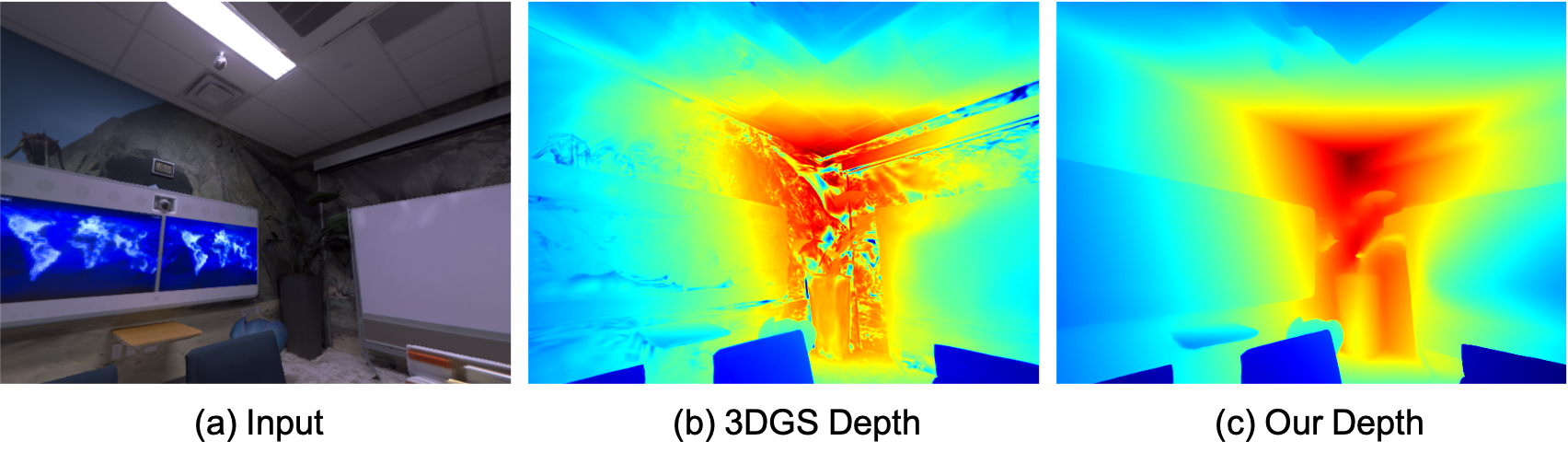}
    \captionsetup{font={footnotesize}}
    \caption{\textbf{Rendered Depth.} The original depth in 3DGS exhibits significant noise, while our depth is smoother and more accurate.}
    \label{fig:depth}
\end{figure}

Although 3DGS achieves high-fidelity novel view rendering and fast training and rendering speeds. As discussed in previous methods~\cite{guedon2023sugar,jiang2023gaussianshader}, Gaussians often do not conform well to actual surfaces, resulting in poor geometric accuracy. Fig.~\ref{fig:depth} also shows this conclusion. Extracting accurate meshes from millions of discrete Gaussian points is an extremely challenging task. The fundamental reason for this lies in the disorderly and irregular nature of Gaussians, which makes them unable to accurately model the surfaces of real scenes. Moreover, optimizing solely based on image reconstruction loss can easily lead to local optima, ultimately resulting in Gaussians failing to conform to actual surfaces and exhibiting poor geometric accuracy. In many practical tasks, geometric reconstruction accuracy is a crucial metric. Therefore, to address these issues, we propose a novel framework based on 3DGS that achieves high-fidelity geometric reconstruction while maintaining the high-quality rendering quality, fast training, and rendering speeds characteristic of 3DGS.

In this paper, we propose a novel unbiased depth rendering method based on 3DGS, facilitating the integration of various geometric constraints to achieve precise geometric estimation. Previous methods ~\cite{jiang2023gaussianshader} render depth by blending the accumulations of each Gaussian at the z-position of the camera, resulting in two main issues as shown in Fig.~\ref{fig:unbias}. The depth corresponds to a curved surface and may deviate from the actual surface. To address these issues, we compress 3D Gaussians into flat planes and blend their accumulations to obtain normal and camera-to-plane distance maps. These maps are then transformed into depth maps. This method involves blending Gaussian plane accumulations to determine a pixel's plane parameters. The intersection of the ray and plane defines the depth, depending on the Gaussian's position and rotation. By dividing the distance map by the normal map, we cancel out the ray accumulation weights, ensuring the depth estimation is unbiased and falls on the estimated plane. In our experiment shown in Fig.~\ref{fig:unbias}, we used true depth to guide two depth rendering methods. After optimization, we mapped the positions of all Gaussian points. Results show that our method produces Gaussians that closely align with the actual surface, while the previous method generates noisy Gaussians that fail to adhere precisely to the surface.

After rendering the plane parameters for each pixel, we apply single-view and multi-view regularization to optimize these parameters. Empirically, adjacent pixels often belong to the same plane. Using this local plane assumption, we compute a normal map from neighboring pixel depth estimations and ensure consistency between this normal map and the rendered normal map. At geometric edges, the local plane assumption fails, so we detect these edges using image edges and reduce the weight in these areas, achieving smooth geometry and consistent depth and normals. However, due to the discrete and unordered nature of Gaussians, geometry may be inconsistent across multiple views. To address this, we apply multi-view regularization ensuring global geometric consistency. Similar to the Eikonal loss~\cite{wang2021neus}, we incorporate a multi-view geometric consistency loss to ensures smooth and consistent geometric reconstruction, even in areas with noise, blur, or weak textures. 

We use two photometric coefficients to compensate for overall changes in image brightness, further improving reconstruction quality. 
Finally, we validate the rendering and reconstruction quality on the MipNeRF360, the DTU~\cite{jensen2014large} and the Tanks and Temples(TnT)~\cite{knapitsch2017tanks} dataset. Experimental results demonstrate that, while maintaining the original Gaussian rendering quality and rendering speed, our method achieves state-of-the-art reconstruction accuracy. Moreover, our training speed only requires one hour on a single GPU, while the state-of-the-art method based on NeRF~\cite{li2023neuralangelo} requires eight GPUs over two days. In summary, our method makes the following contributions:

\begin{itemize}
  \item We propose ~\textbf{a novel unbiased depth rendering method}. Based on this rendering method, we can render the reliable plane parameters for each pixel, facilitating the incorporation of various geometric constraints.
  \item We introduce ~\textbf{single-view and multi-view regularizations} to optimize the plane parameters of each pixel, achieving high-precision global geometric consistency.
  \item \textbf{The exposure compensation} simply and effectively enhances reconstruction accuracy. 
  \item Our method, while maintaining the high rendering accuracy and speed of the original GS, achieves \textbf{state-of-the-art reconstruction accuracy}, and our training time is near \textbf{100 times faster} compared to state-of-the-art reconstruction methods based on NeRF~\cite{li2023neuralangelo}.
\end{itemize}

\begin{figure*}[htb]
    \centering
    \includegraphics[width=0.98\linewidth]{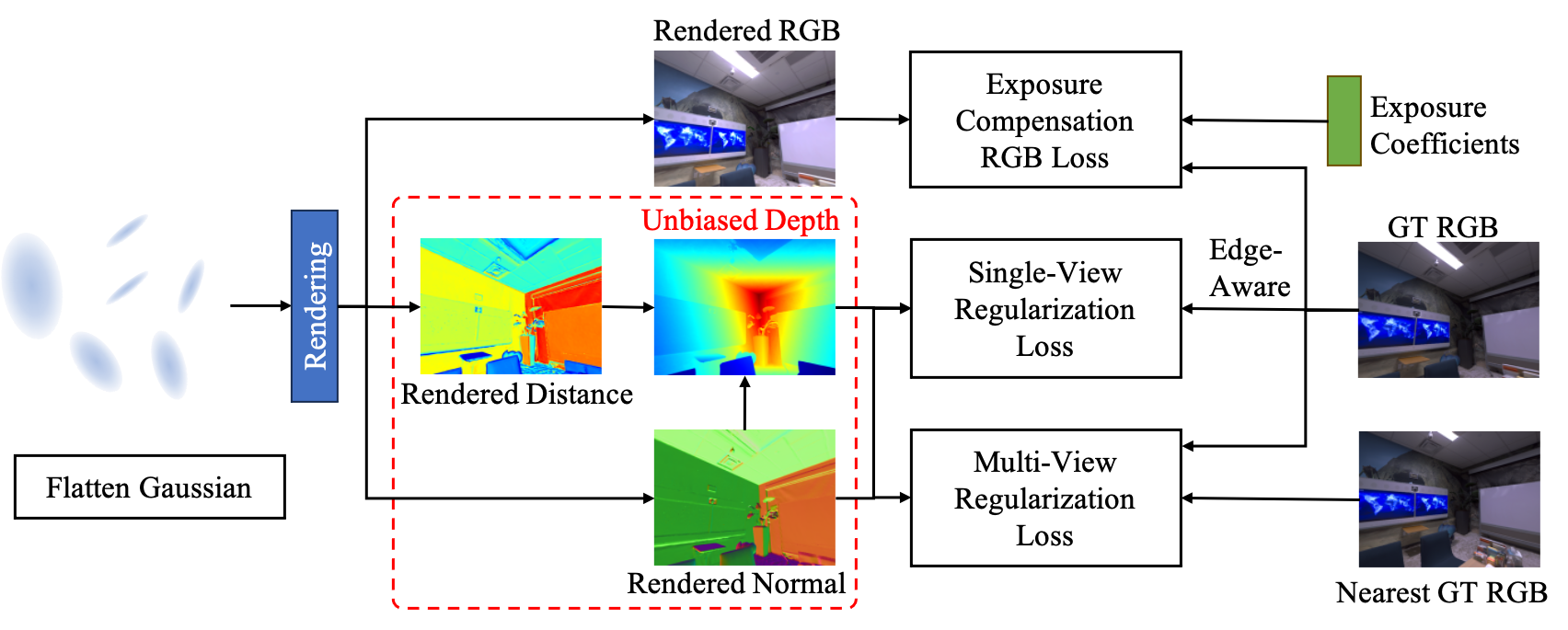}
    \captionsetup{font={footnotesize}}
    \caption{\textbf{PGSR Overview.} We compress Gaussians into flat planes and render distance and normal maps, which are then transformed into unbiased depth maps. Single-view and multi-view geometric regularization ensure high precision in global geometry. Exposure compensation RGB loss enhances reconstruction accuracy. 
    }
    \label{fig:system}
    \vspace{-0.3cm}
\end{figure*}

\section{Related Work}
Surface reconstruction is a cornerstone field in computer graphics and computer vision, aimed at generating intricate and accurate surface representations from sparse or noisy input data. Obtaining high-fidelity 3D models from real-world environments is pivotal for enabling immersive experiences in augmented reality (AR) and virtual reality (VR). This paper focuses exclusively on surface reconstruction under given poses, which can be readily computed using SLAM~\cite{campos2021orb, chen2021rnin,chen2022vip,matsuki2023gaussian} or SFM~\cite{schonberger2016structure, Moulon2012, wu2013towards} methods.

\subsection{Traditional Surface Reconstruction}

Traditional methods adhere to the universal multi-view stereo pipeline, which can be roughly categorized based on the intermediate representation they rely on, such as point cloud~\cite{lhuillier2005quasi, Accurate}, volume~\cite{kutulakos2000theory}, depth map~\cite{schoenberger2016mvs, campbell2008using, galliani2015massively}, etc. The commonly used method separates the 
overall MVS problem into several parts, by initially extracting dense point clouds from multi-view images through block-based matching~\cite{barnes2009patchmatch}, followed by the construction of surface structures either through triangulation~\cite{cazals2006delaunay} or implicit surface fitting~\cite{kazhdan2006poisson, kazhdan2013screened}. Despite being well-established and extensively utilized in academia and industry, these traditional methods are susceptible to artifacts stemming from erroneous matching or noise introduced during the pipeline. In response, several approaches aim to enhance reconstruction completeness and accuracy by integrating deep neural networks into the matching process \cite{wang2021patchmatchnet, sarlin2019coarse}.

\subsection{Neural Surface Reconstruction}

Numerous pioneering efforts have leveraged pure deep neural networks to predict surface models directly from single or multiple image conditions using point clouds~\cite{Fan:2017:Point, Lin:2018:Learning}, voxels\cite{Choy:2016:3d, Xie:2019:Pix2vox}, and triangular meshes~\cite{Wang:2018:Pixel2mesh, li2020saliency} or implicit fields~\cite{Park:2019:Deepsdf, Mescheder:2019:Occupancy} in end-to-end manner. However, these methods often incur significant computational overhead during network inference and demand extensively labeled training 3D models, hindering their real-time and real-world applicability.


With the rapid advancement in neural surface reconstruction tasks, a meticulously designed scene recovery method named NeRF~\cite{mildenhall2021nerf} emerged. NeRF-based methods take 5D ray information as input and predict density and color sampled in continuous space, yielding notably more realistic rendering results. However, this representation falls short in capturing high-fidelity surfaces.

Consequently, several approaches have transformed NeRF-based network architectures into surface reconstruction frameworks by incorporating intermediate representations such as occupancy~\cite{Niemeyer:2020:Differentiable} or signed distance fields~\cite{wang2021neus, Yariv:2021:Volume}. Despite the potent surface reconstruction capabilities exhibited by NeRF-based frameworks, the stacked multi-layer-perceptron (MLP) layers impose constraints on inference time and representation ability. To address this challenge, various following studies aim to reduce dependency on MLP layers by decomposing scene information into separable structures, such as points~\cite{xu2022point} and voxels~\cite{Liu:2020:Neural, li2022vox, li2023neuralangelo}.


\subsection{Gaussian Splatting based Surface Reconstruction}
SuGaR~\cite{guedon2023sugar} proposed a method to extract Mesh from 3DGS. They introduced regularization terms to encourage Gaussian fitting to the scene surface. By sampling 3D point clouds from the Gaussian using the density field, they utilized Poisson reconstruction to extract a mesh from these sampled point clouds. 
\textcolor{r}{
However, biased depth is used to constrain the density field, with the aim of extracting surface points from the density field. The final surface quality depends on the depth quality, and it is difficult to reconstruct smooth surfaces from a discrete density field. Due to the discreteness and randomness of Gaussian points, relying solely on image reconstruction constraints without proper geometric regularization can easily result in local optimization, making it difficult to reconstruct high-precision surfaces. While our method shares some conceptual similarities with SuGaR, such as approximating 3D Gaussian ellipsoids as planes, using the shortest axis as the plane normal representation, and aiming to represent actual surfaces with planes, there are significant differences in the plane rendering method and the use of planes. The concurrent works that are very close in time to ours are 2DGS~\cite{huang20242d} and GOF~\cite{yu2024gaussian}.}
2DGS achieves consistent geometry across views by collapsing the 3D volume into a collection of 2D oriented planar Gaussian disks. GOF forms a Gaussian opacity field, facilitating geometry extraction by directly identifying its level set. \textcolor{r}{However, these Gaussian splatting-based methods still fail to produce high-precision depth and cannot ensure multi-view geometric consistency. 2DGS uses planes to resolve the 3D Gaussian geometric ambiguity in multi-view scenarios. It uses two depth rendering methods, requiring manual selection between the median and expected depth value of ray-plane intersections. In boundary scenarios, 2DGS recommends using median depth. However, median depth suffers from the issue of 'disk-aliasing'. Additionally, there are no constraints to ensure multi-view consistency.} To address these issues, we flattened the Gaussian into a planar shape, which is more suitable for modeling actual surfaces and facilitates rendering parameters such as normals and distances from the plane to the origin. Based on these plane parameters, we proposed unbiased depth estimation, allowing us to extract geometric parameters from the Gaussian. Then, we introduced geometric regularization terms from single-view and multi-view to optimize these geometric parameters, achieving globally consistent high-precision geometric reconstruction.

\section{Preliminary of 3D Gaussian Splatting}
3DGS~\cite{kerbl20233d} explicitly represents 3D scenes with a set of 3D Gaussians $\{\mathcal G_i\}$. Each Gaussian is defined by a Gaussian function:
$$
\mathcal G_i(\bm{x}|\bm{\mu}_i, \bm{\Sigma_i}) = e^{-\frac{1}{2}(\bm{x} - \bm{\mu}_i)^\top\bm{\Sigma}_i^{-1}(\bm{x}-\bm{\mu}_i)},
$$
where $\bm{\mu}_i \in \mathbb R^3$ and $\bm{\Sigma}_i \in \mathbb R^{3\times3}$ are the center of a point $\bm{p}_i\in \mathcal P$ and corresponding 3D covariance matrix, respectively. The covariance matrix $\bm{\Sigma}_i$ can be decomposed into a scaling matrix $\bm{S}_i\in \mathbb R^{3\times3}$ and a rotation matrix $\bm{R}_i\in \mathbb R^{3\times3}$:
$$
\bm{\Sigma}_i = \bm{R}_i\bm{S}_i\bm{S}_i^\top \bm{R}_i^\top.
$$

3DGS allows fast $\alpha$-blending for rendering. Given a transformation matrix $W$ and an intrinsic matrix $\bm{K}$, $\bm{\mu_i}$ and $\bm{\Sigma}_i$ can be transformed to camera coordinate corresponding to $\bm{W}$ and then projected to 2D coordinate:
$$
\bm{\mu}_i^{'}=\bm{KW}[\bm{\mu}_i,1]^\top, \quad  \bm{\Sigma}_i^{'}=\bm{JW\Sigma}_i\bm{W}^\top \bm{J}^\top,
$$
where $J$ denotes the Jacobian matrix of the projective transformation. Rendering color $\bm{C}\in\mathbb R^3$ of a pixel $\bm u$ can be obtained in a manner of $\alpha$-blending:
$$
\bm{C} = \sum_{i\in N} T_i \alpha_i \bm{c}_i,\quad T_i=\prod_{j=1}^{i-1}(1 - \alpha_i),
$$
where $\alpha_i$ is calculated by evaluating $\mathcal G_i(\bm{u}|\bm{\mu}_i^{'}, \bm{\Sigma}_i^{'})$ multiplied with a learnable opacity corresponding to $\mathcal G_i$, and the view-dependent color $\bm{c}_i\in\mathbb R^3$ is represented by spherical harmonics (SH) from the Gaussian $\mathcal G_i$. $T_i$ is the cumulative opacity. $N$ is the number of Gaussians that the ray passes through. 

The center $\bm{\mu}_i$ of a Gaussian $\mathcal G_i$. can be projected into the camera coordinate system as:
$$
\begin{bmatrix}
    x_i,y_i,z_i,1
\end{bmatrix}^\top=\bm{W}[\bm{\mu}_i,1]^\top,
$$
Previous Methods\cite{jiang2023gaussianshader,cheng2024gaussianpro} render depth under the current viewpoint:
$$
\bm{D} = \sum_{i\in N} T_i \alpha_i z_i.
\label{eq:render_depth}
$$

\begin{figure*}[htbp]
    \centering
    \includegraphics[width=0.98\linewidth]{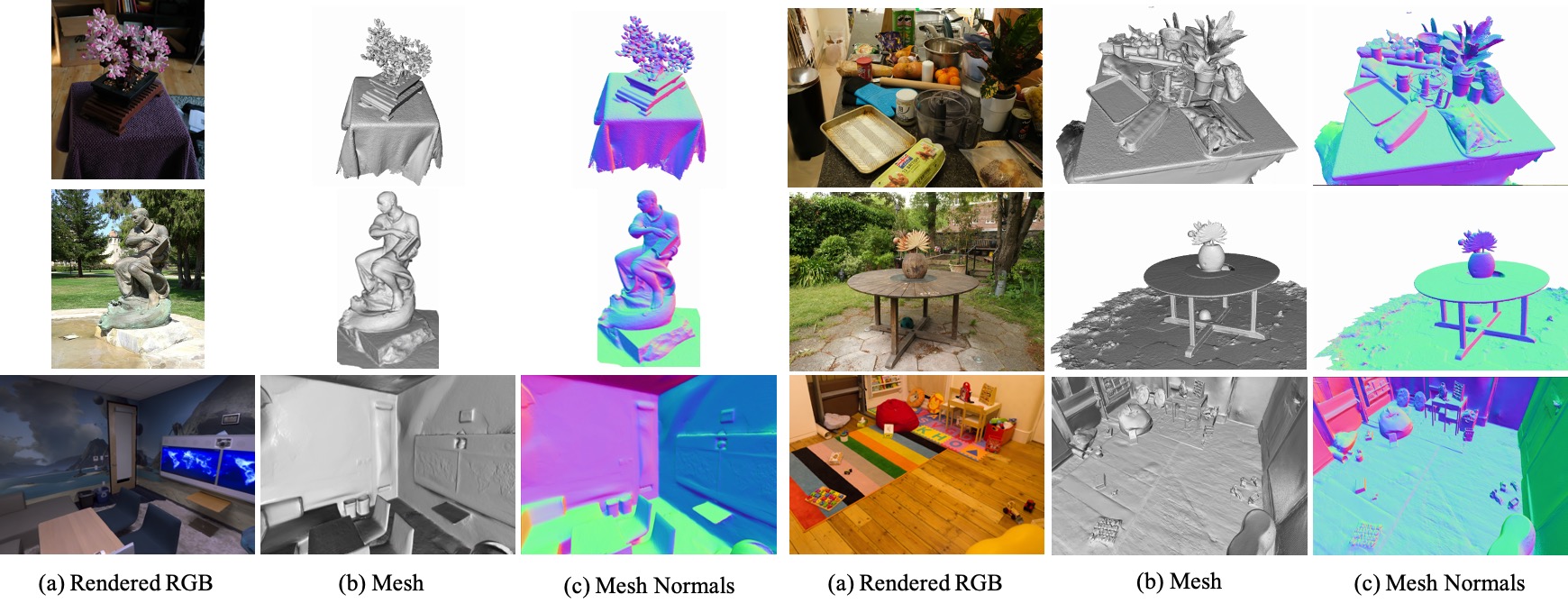}
    \captionsetup{font={footnotesize}}
    \caption{\textbf{The rendering and mesh reconstruction results in various indoor and outdoor scenes that we have achieved.} PGSR achieves high-precision geometric reconstruction from a series of RGB images without requiring any prior knowledge.}
    \label{fig:reconstruction2}
    \vspace{-0.3cm}
\end{figure*}

\section{Method}
Given multi-view RGB images of static scenes, our goal is to achieve efficient and high-fidelity scene geometry reconstruction and rendering quality. Compared to 3DGS, we achieve global consistency in geometry reconstruction while maintaining similar rendering quality. Initially, we improve the modeling of scene geometry attributes by compressing 3D Gaussians into a 2D flat plane representation, which is used to generate plane distance and normal maps, and subsequently converted into unbiased depth maps. We then introduce single-view geometric, multi-view photometric, and geometric consistency loss to ensure global geometry consistency. Additionally, the exposure compensation model further improves reconstruction accuracy. 

\begin{figure}[htb]
    \centering
    \includegraphics[width=0.6\linewidth]{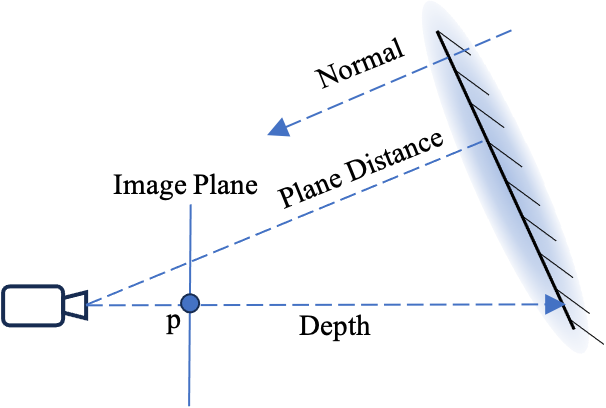}
    \captionsetup{font={footnotesize}}
    \caption{\textbf{Unbiased Depth.}}
    \label{fig:depth2}
    \vspace{-0.3cm}
\end{figure}

\subsection{Planar-based Gaussian Splatting Representation}
In this section, we will discuss how to transform 3D Gaussians into a 2D flat plane representation. Based on this plane representation, we introduce an unbiased depth rendering method, which will render plane-to-camera distance and normal maps, and can then be converted into depth maps. With geometric depth, distance, and normal maps available, it becomes easier to introduce single-view regularization and multi-view regularization in the following sections.

Due to the difficulty in modeling real-world scene geometry attributes such as depth and normals using 3D Gaussian shapes, it's necessary to flatten the 3D Gaussians into 2D flat Gaussians in order to accurately represent the geometry surface of the actual scene. Achieving precise geometry reconstruction and high-quality rendering requires the 2D flat Gaussians to accurately conform to the scene surface. Since the 2D flat Gaussians approximate a local plane, we can conveniently render the depth and normals of the scene.

\textbf{Flattening 3D Gaussian:}
The covariance matrix $\bm{\sum}_i=\bm{R}_i\bm{S}_i\bm{S}_i^T\bm{R}_i^T$ of a 3D Gaussian expresses the ellipsoidal shape. Here, $\bm{R}_i$ represents the orthonormal basis of the ellipsoid's three axes, and the scale factor $\bm{S}_i$ defines the size along each direction. By compressing the scale factor along specific axes, the Gaussian ellipsoid can be flattened into planes aligned with those axes. We compress the Gaussian ellipsoid along the direction of the minimum scale factor, effectively flattening the ellipsoid into a plane closest to its original shape. According to the method\cite{chen2023neusg}, we directly minimize the minimum scale factor $\bm{S}_i=diag(s_1, s_2, s_3)$ for each Gaussian:
\begin{equation}
    L_s=\parallel \min(s_1, s_2, s_3) \parallel_1.
\end{equation}

\textbf{Unbiased Depth Rendering:}
\label{sec.Unbiased Depth Rendering} 
The direction of the minimum scale factor corresponds to the normal $\bm{n}_i$ of the Gaussian. Due to the ambiguity of the normal direction when there are two directions for the shortest axis, we resolve this issue by using the viewing direction to determine the normal direction. This implies that the angle between the viewing direction and the normal direction should be greater than 90 degrees. The final normal map under the current viewpoint is achieved through $\alpha$-blending:
\begin{equation}
    \bm{N}=\sum_{i\in N}\bm{R}_c^T\bm{n}_i\alpha_i \prod_{j=1}^{i-1}(1-\alpha_j),
\label{eq:render_normal}
\end{equation}
where $\bm{R}_c$ is the rotation from the camera to the global world. 
The distance from the plane to the camera center can be expressed as \textcolor{r}{$\textit{d}_i=(\bm{R}_c^T(\bm{\mu}_i-\bm{T}_c))^T(\bm{R}_c^T\bm{n}_i)$}, where $\bm{T}_c$ is the camera center in the world. $\bm{\mu}_i$ is the center of gaussian $G_i$. The final distance map under the current viewpoint is achieved through $\alpha$-blending:
\begin{equation}
    \bm{\mathcal{D}}=\sum_{i\in N}d_i\alpha_i \prod_{j=1}^{i-1}(1-\alpha_j).
\end{equation}
Referencing Fig.~\ref{fig:depth2}, after obtaining the distance and normal of the plane through rendering, we can determine the corresponding depth map by intersecting rays with the plane:
\begin{equation}
    \bm{D}(\bm{p})=\frac{\bm{\mathcal{D}}}{\bm{N}(p)\bm{K}^{-1}\tilde{\bm{p}}},
\end{equation}
where $\bm{p}=[u,v]^T$ indicates the 2D position on the image plane. $\tilde{\bm{p}}$ is the homogeneous coordinate representation of $\bm{p}$, and $\bm{K}$ refers to the intrinsic of camera.

As shown in Fig.~\ref{fig:unbias}, our method of rendering depth has two major advantages compared to other depth rendering techniques. First, Our depth shapes are consistent with flattened Gaussian shapes, which can truly reflect actual surfaces. Previous methods typically involve directly rendering the depth map based on $\alpha$-blending of the depth Z of Gaussians. Their depth is curved, inconsistent with the flat Gaussian shape, causing geometric conflicts. In contrast, we render the normal and distance maps of the plane first and then convert them into the depth map. Our depth lies on the Gaussian \textcolor{r}{flat plane}. When the 3D Gaussian flat planes fit the actual surface, the rendered depth can ensure complete consistency with the actual surface. Second, since the accumulation weight for each ray may be less than 1, previous rendering methods are affected by the weight accumulation, potentially resulting in depths that are closer to the camera side and overall underestimated. In contrast, our depth is obtained by dividing the distance from the rendering origin to the plane by the normal, effectively eliminating the influence of weight accumulation coefficients.

\begin{figure}[htb]
    \centering
    \includegraphics[width=1\linewidth]{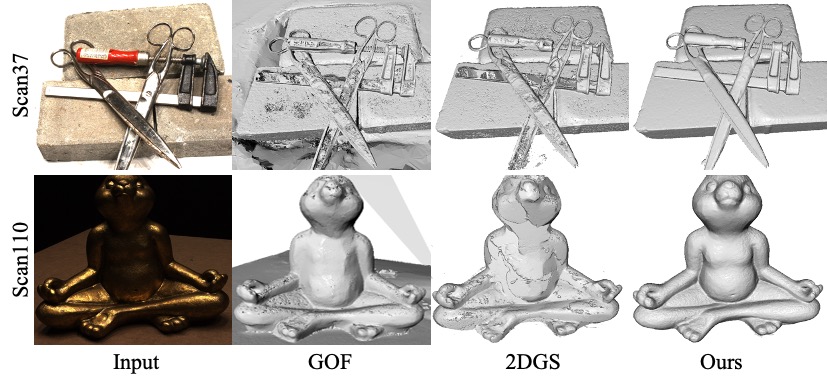}
    \captionsetup{font={footnotesize}}
    \caption{\textbf{Qualitative comparison on DTU dataset.} PGSR produces smooth and detailed surfaces.}
    \label{fig:reconstruction_dtu}
    \vspace{-0.3cm}
\end{figure}

\begin{figure*}[htb]
    \centering
    \includegraphics[width=0.98\linewidth]{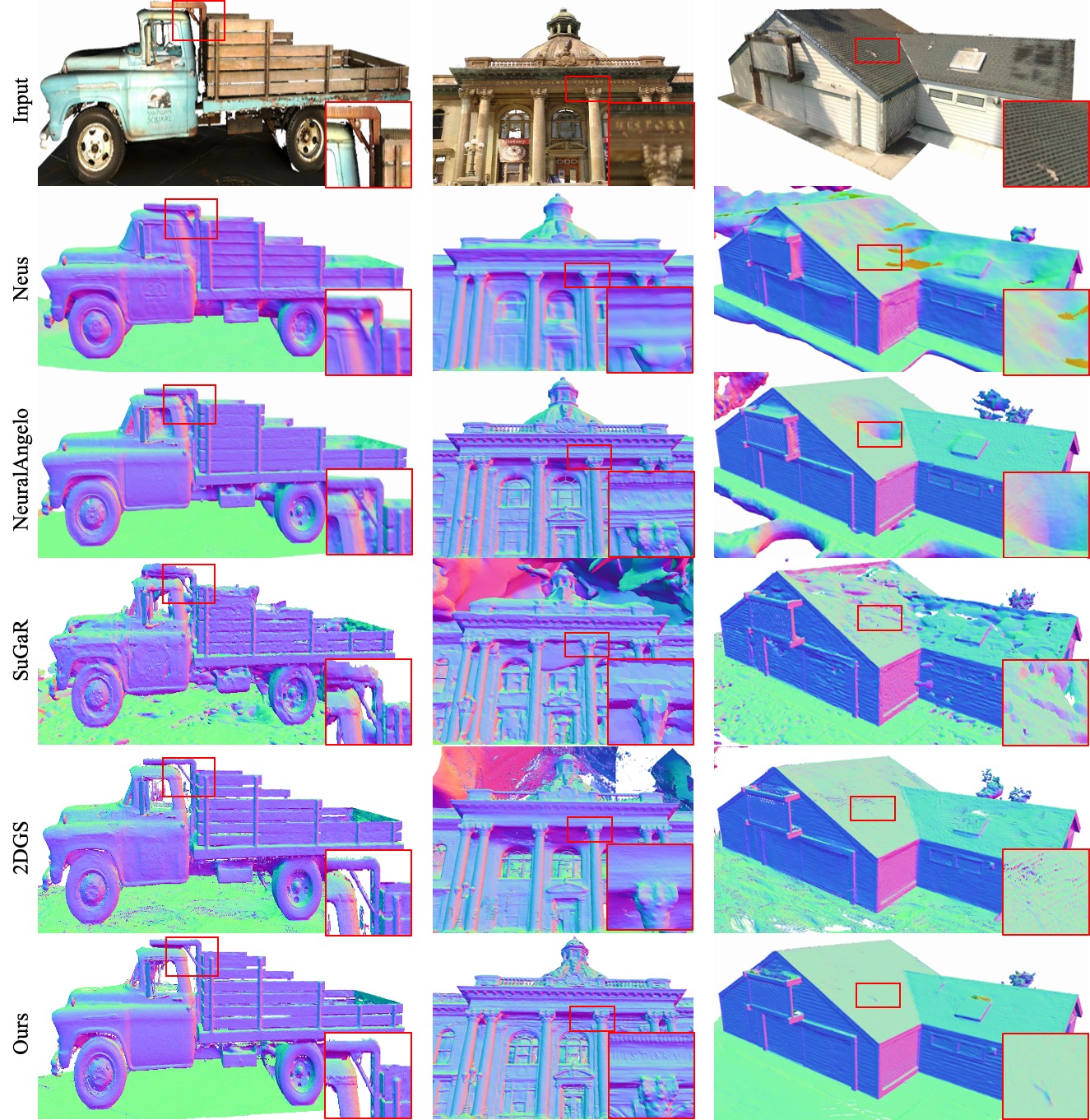}
    \captionsetup{font={footnotesize}}
    \caption{\textbf{Qualitative comparison on Tanks and Temples dataset.} We visualize surface quality using a normal map generated from the reconstructed mesh. PGSR outperforms other baseline approaches in capturing scene details, whereas baseline methods exhibit missing or noisy surfaces.}
    \label{fig:reconstruction1}
    \vspace{-0.3cm}
\end{figure*}

\subsection{Geometric Regularization}

\subsubsection{Single-View Regularization}
The original 3DGS relying solely on image reconstruction loss can easily fall into local overfitting optimization, leading to Gaussian shapes inconsistent with the actual surface. Based on this, we introduce geometric constraints to ensure that the 3D Gaussian fits the actual surface as closely as possible.

\textbf{Local Plane Assumption:}
Encouraged by these methods~\cite{qi2018geonet,jiang2023gaussianshader,long2024adaptive}, we adopt the assumption of local planarity to constrain the local consistency of depth and normals, meaning a pixel and its neighboring pixels can be considered as an approximate plane. After rendering the depth map, we sample four neighboring points using a fixed template. With these known depths, we compute the plane's normal. This process is repeated for the entire image, generating normals from the rendered depth map. We then minimize the difference between this normal map and the rendered normal map, ensuring geometric consistency between local depth and normals.

\textbf{Image Edge-Aware Single-View Loss:}
Neighboring pixels may not necessarily fully adhere to the local planarity assumption, especially in edge regions. To address this issue, We use image edges to approximate geometric edges. For a pixel point $\bm{p}$, we sample four points from the neighboring pixels, such as up, down, left, and right. We project the four sampled depth points into 3D points $\{\bm{P}_j|j=1,...,4\}$ in the camera coordinate system, then calculate the normal of the local plane for the pixel point $\bm{p}$ is:
\begin{equation}
    \bm{N}_d(\bm{p})=\frac{(\bm{P}_1-\bm{P}_0)\times(\bm{P}_3-\bm{P}_2)}{|(\bm{P}_1-\bm{P}_0)\times(\bm{P}_3-\bm{P}_2)|},
\end{equation}
Finally, we add the single-view normal loss is:
\begin{equation}\color{r}
    \bm{L}_{svgeom}=\frac{1}{W}\sum_{\bm{p}\in W}(1-\overline{\nabla \bm{I}} )^2 \parallel \bm{N}_d(\bm{p}) - \bm{N}(\bm{p}) \parallel_1,
\end{equation}
where $\overline{\nabla \bm{I}}$ is the image gradient normalized to the range of 0 to 1, $\bm{N}(\bm{p})$ is from Equation~\ref{eq:render_normal}, and $W$ is the set of image pixels.

\subsubsection{Multi-View Regularization}
Single-view geometry regularization can maintain consistency between depth and normal geometry, providing fairly accurate initial geometric information. However, due to the irregular discretization of Gaussian point cloud optimization, we found that the geometry structure across multiple views is not entirely consistent. Therefore, it is necessary to introduce multi-view geometry regularization to ensure global consistency of the geometry structure.

\begin{figure}[t]
    \centering
    \includegraphics[width=0.9\linewidth]{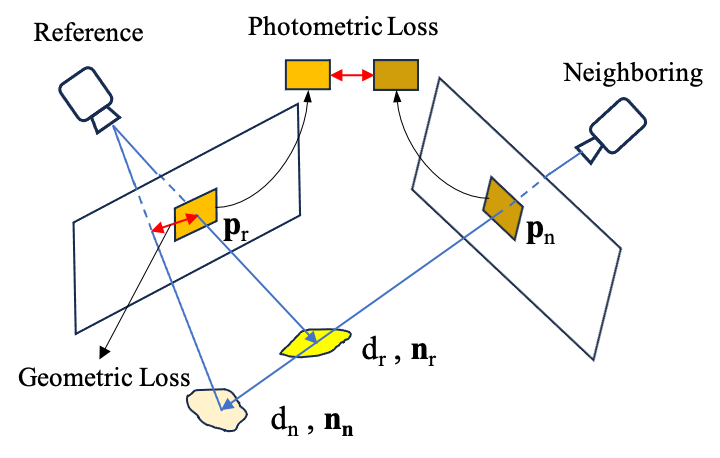}
    \captionsetup{font={footnotesize}}
    \caption{
    \textbf{Multi-view photometric and geometric loss.}
    }
    \label{fig:multi_view}
    \vspace{-0.3cm}
\end{figure}

\textbf{Multi-View Geometric Consistency:}
The image loss often suffers from influences such as image noise, blur, and weak textures. In these cases, the geometric solution for photometric consistency is unreliable. Due to the discrete nature of Gaussian properties, we cannot establish a spatially dense or semi-dense SDF field as in SDF methods based on NeRF. We are unable to use spatial smoothness constraints, such as the Eikonal loss~\cite{wang2021neus}, to avoid the influence of unreliable solutions. To mitigate the impact of unreliable geometric solutions and ensure multi-view geometric consistency, we introduce this consistency prior constraint, which helps converge to the correct solution position, enhancing geometric smoothness. 

We render the normals $ \bm{N} $ and the plane distances $ \bm{\mathcal{D}}$ to the camera for both the reference frame and the neighboring frame. As shown in Fig.~\ref{fig:multi_view}, for a specific pixel $ \bm{p}_r $ in the reference frame, the corresponding normal is $ \bm{n}_r $ and the distance is $ d_r $. The pixel $ \bm{p}_r $ in the reference frame can be mapped to a pixel $ \bm{p}_n $ in the neighboring frame through the homography matrix $\bm{H}_{rn}$:
\begin{equation}
\tilde{\bm{p}_n}=\bm{H}_{rn}\tilde{\bm{p}_r},
\end{equation}
\begin{equation}
\bm{H}_{rn}=\bm{K}_n(\bm{R}_{rn}-\frac{\bm{T}_{rn}\bm{n}_r^T}{d_r})\bm{K}_r^{-1},
\end{equation}
where $\bm{R}_{rn}$ and $\bm{T}_{rn}$ are the relative rotation and translation from the reference frame to the neighboring frame. Similarly, for the pixel $ \bm{p}_n $ in the neighboring frame, we can obtain the normal $ \bm{n}_n $ and the distance $ d_n $ to compute the homography matrix $ \bm{H}_{nr} $. The pixel $ \bm{p}_r $ undergo forward and backward projections between the reference frame and the neighboring frame through $ \bm{H}_{rn} $ and $ \bm{H}_{nr} $. Minimizing the forward and backward projection error constitutes the multi-view geometric consistency regularization:
\begin{equation}\color{r}
    \bm{L}_{mvgeom}=\frac{1}{V}\sum_{\bm{p}_r\in W}w(\bm{p}_r)\phi(\bm{p}_r),
\end{equation}
\begin{equation}\color{r}
   w(\bm{p}_r)=\begin{cases}1/exp(\phi(\bm{p}_r)),
 & \text{ if } \phi(\bm{p}_r) < 1
 \\0,
  & \text{ if } \phi(\bm{p}_r) >= 1
\end{cases},
\end{equation}
where 
$
    \phi(\bm{p}_r)=\parallel \bm{p}_r-\bm{H}_{nr}\bm{H}_{rn}\bm{p}_r \parallel
$ is the forward and backward projection error of $\bm{p}_r$. When $\phi(\bm{p}_r)$ exceeds a certain threshold, it can be considered that the pixel is occluded or that there is a significant geometric error. To prevent errors caused by occlusion, these pixels will not be included in the multi-view regularization term. If these pixels are mistakenly identified as occluded due to geometric errors, it does not affect our final convergence. This is because the single-view regularization term and the use of sparse 3D Gaussians to represent dense scenes will gradually propagate high-precision geometry, eventually leading all Gaussians to converge to the correct positions. \textcolor{r}{$w(\bm{p}_r)$ is a weight of geometric occlusion estimation, and the larger the projection error, the smaller the weight. During training, the gradient of the weight will be detached.}

\textbf{Multi-View Photometric Consistency:}
Drawing inspiration from multi-view Stereo (MVS methods)~\cite{schonberger2016structure,campbell2008using,fu2022geo}, we employ photometric multi-view consistency constraints based on plane patches. We map a $7\times 7$ pixel patch $ \bm{P}_r $ centered at $ \bm{p}_r $ to the neighboring frame patch $ \bm{P}_n $ using the homography matrix $ \bm{H}_{rn} $. Focusing on geometric details, we convert color images into grayscale. Multi-view photometric regularization requires that $ \bm{P}_r $ and $ \bm{P}_n $ should be as consistent as possible. We use the normalized cross correlation (NCC)\cite{yoo2009fast} of patches in the reference frame and the neighboring frame to measure the photometric consistency:
\begin{equation}\color{r}
    \bm{L}_{mvrgb}=\frac{1}{V}\sum_{\bm{p}_r\in W}w(\bm{p}_r)(1-NCC(\bm{I}_r(\bm{p}_r), \bm{I}_n(\bm{H}_{rn}\bm{p}_r))),
\end{equation}

\subsubsection{Geometric Regularization Loss}

Finally, the geometric regularization loss includes single-view geometric, multi-view geometric, and multi-view photometric consistency constraints:

\begin{equation}
    \bm{L}_{geom}=\lambda_2\bm{L}_{svgeom}+\lambda_3\bm{L}_{mvrgb}+\lambda_4\bm{L}_{mvgeom}.
\end{equation}

\begin{table*}\color{r}
\captionsetup{font={footnotesize}}
  \caption{\textbf{Quantitative results of rendering quality for novel view synthesis on Mip-NeRF360 dataset.} "Red", "Orange" and "Yellow" denote the best, second-best, and third-best results. PGSR achieves results close to 3DGS and outperforms similar reconstruction method SuGaR.}
  \label{tab:view_eval}
  \scriptsize%
  \centering%
  \setlength{\tabcolsep}{10pt}
  \scalebox{1.0}[1.0]{
  \begin{tabular}{c|c|ccc|ccc|ccc}
  \hline
    \multicolumn{2}{c}{} &\multicolumn{3}{c}{Indoor scenes} &\multicolumn{3}{c}{Outdoor scenes}&\multicolumn{3}{c}{Average on all scenes} \\
   \multicolumn{2}{c}{} & PSNR$\uparrow$ & SSIM$\uparrow$ & LPIPS$\downarrow$ & PSNR$\uparrow$ & SSIM$\uparrow$ & LPIPS$\downarrow$ & PSNR$\uparrow$ & SSIM$\uparrow$ & LPIPS$\downarrow$ \\
  \hline
  \multirow{5}{*}{\rotatebox{90}{NeRF-based}} &
  NeRF~\cite{mildenhall2021nerf} & 26.84 & 0.790 & 0.370 & 21.46  & 0.458  & 0.515  & 24.15  & 0.624 & 0.443 \\
  & Deep Blending~\cite{hedman2018deep} & 26.40 & 0.844 & 0.261 & 21.54 & 0.524 & 0.364 & 23.97 & 0.684 & 0.313 \\
  & INGP~\cite{muller2022instant} & 29.15 & 0.880 & 0.216 & 22.90 & 0.566  & 0.371  & 26.03  & 0.723 & 0.294  \\
  & M-NeRF360~\cite{barron2021mip} & \best{31.72} & 0.917 & 0.180 & \sbest24.47  & 0.691  & \tbest0.283  & \best28.10  & \tbest0.804 & \tbest0.232 \\
&  Neus~\cite{wang2021neus} & 25.10 & 0.789 & 0.319 & 21.93 & 0.629  & 0.600 & 23.74  & 0.720 & 0.439 \\
  \midrule
  \multirow{3}{*}{\rotatebox{90}{\small GS-based}} &
  3DGS~\cite{kerbl20233d} & \sbest{30.99} & \tbest0.926 & 0.199 & 24.24  &  0.705  & \tbest0.283  & 27.24  & 0.803 & 0.246 \\
 & SuGaR~\cite{guedon2023sugar} & 29.44 & 0.911 & 0.216 & 22.76  & 0.631  & 0.349  & 26.10  & 0.771 & 0.283 \\
 & 2DGS~\cite{huang20242d} & 30.39 & 0.923 & \tbest 0.183 & \tbest 24.33  & \tbest 0.709  & 0.284  & 27.03  & \tbest 0.804 &  0.239 \\
 & GOF~\cite{yu2024gaussian} & \tbest30.80 & \sbest 0.928 & \sbest 0.167 & \best 24.76  & \sbest 0.742  & \sbest 0.225  & \sbest 27.78  & \best 0.835 & \sbest 0.196 \\
 & PGSR & 30.36 & \best0.934 & \best 0.147 & \best24.76 & \best0.752 & \best0.203 & \tbest 27.25 & \sbest 0.833 & \best0.178 \\
  \bottomrule
  \end{tabular}%
  }
\vspace{-0.2cm}
\end{table*}

\subsection{Exposure Compensation Image Loss}
Due to changes in external lighting conditions, cameras may have different exposure times during different shooting moments, leading to overall brightness variations in images. The original 3DGS does not consider brightness changes, which can result in floating artifacts in practical scenes. To model the overall brightness variations at different times, we assign two exposure coefficients, $a$ and $b$, to each image. Ultimately, images with exposure compensation can be obtained by simply computing with exposure coefficients:
\begin{equation}
    \bm{I}_i^a=exp(a_i)\bm{I}_i^r+b_i,
\end{equation}
where $\bm{I}_i^r$ is the rendered image and $\bm{I}_i^a$ is the exposure-adjusted image.
We employ the following image loss:
\begin{equation}
    \bm{L}_{rgb}=(1-\lambda)\bm{L}_1(\tilde{\bm{I}}-\bm{I}_i)+\lambda\bm{L}_{SSIM}(\bm{I}_i^r-\bm{I}_i).
\end{equation}
\begin{equation}
    \tilde{\bm{I}} =\begin{cases}\bm{I}_i^a,
 & \text{ if } \bm{L}_{SSIM}(\bm{I}_i^r-\bm{I}_i) < 0.5
 \\\bm{I}_i^r,
  & \text{ if } \bm{L}_{SSIM}(\bm{I}_i^r-\bm{I}_i) >= 0.5
\end{cases}
\end{equation}
where $\bm{I}_i$ is the ground truth image. The L1 loss constraint ensures that the exposure-adjusted image is consistent with the ground truth image, while the SSIM loss requires the rendered image to have similar structures to the ground truth image. To enhance the robustness of exposure coefficient estimation, we need to ensure that the rendered image and the ground truth image have sufficient structural similarity before performing the estimation. After training, $\bm{I}_i^r$ is required to be globally consistent and maintain structural similarity with the ground truth image, while $\bm{I}_i^a$ can adjust the brightness of images to match the ground truth image perfectly.

\subsection{Training}
In summary, our final training loss $\bm{L}$ consists of the image reconstruction loss $\bm{L}_{rgb}$, the flattening 3D Gaussian loss $\bm{L}_{s}$, the geometric loss $\bm{L}_{geom}$:
\begin{equation}
    \bm{L}=\bm{L}_{rgb}+\lambda_1\bm{L}_{s}+\bm{L}_{geom}.
\end{equation}
We set $\lambda_1=100$. For the image reconstruction loss, we set $\lambda=0.2$. For the geometric loss, we set $\lambda_2=0.015$, $\lambda_3=0.15$, and  $\lambda_4=0.03$.

\setlength\tabcolsep{0.5em}
\begin{table*}[tb]\color{r}
\captionsetup{font={footnotesize}}
\caption{\textbf{Quantitative results of chamfer distance(mm)$\downarrow$ on DTU dataset\cite{jensen2014large}.} PGSR achieves the highest reconstruction accuracy and is over 100 times faster than the SDF method based on NeRF.} 
\centering
\resizebox{0.95\textwidth}{!}{%
\begin{tabular}{@{}lccccccccccccccccclc|c}
\toprule
 \multicolumn{3}{c}{} & 24 & 37 & 40 & 55 & 63 & 65 & 69 & 83 & 97 & 105 & 106 & 110 & 114 & 118 & 122 & & Mean & Time \\ \midrule
 & VolSDF\cite{Yariv:2021:Volume} & & 1.14 & 1.26 & 0.81 & 0.49 & 1.25 & 0.70 & 0.72 & \tbest1.29 & 1.18 & \tbest0.70 & 0.66 & 1.08 & 0.42 & 0.61 & 0.55 & & 0.86 & \texttt{>} 12h\\
 & NeuS\cite{wang2021neus} & & 1.00 & 1.37 & 0.93 & 0.43 & 1.10 & \tbest 0.65 &  \tbest0.57 & 1.48 & \tbest1.09 & 0.83 & \tbest0.52 & 1.20 & \tbest0.35 & \tbest0.49 & 0.54 & & 0.84 & \texttt{>} 12h\\
 & Neuralangelo\cite{li2023neuralangelo} & & \sbest 0.37 & \sbest0.72 & \best 0.35 & \sbest 0.35 & \sbest 0.87 & \best 0.54 & \sbest 0.53 & \tbest1.29 & \sbest 0.97 & 0.73 & \sbest 0.47 & \sbest 0.74 & \sbest 0.32 & \sbest 0.41 &  \sbest 0.43 & & \sbest 0.61 & \texttt{>} 128h\\ \midrule
 & SuGaR\cite{guedon2023sugar} && 1.47 & 1.33 & 1.13 & 0.61 & 2.25 & 1.71 & 1.15 & 1.63 & 1.62 & 1.07 & 0.79 & 2.45 & 0.98 & 0.88 & 0.79 & & 1.33 & \tbest1h\\
 & 2DGS\cite{huang20242d} && 0.48 & 0.91 & 0.39 & 0.39 & \tbest1.01 & 0.83 & 0.81 & 1.36 & 1.27 & 0.76 & 0.70 & 1.40 & 0.40 & 0.76 & 0.52 & & 0.80 & \best 0.32h\\
  & GOF~\cite{yu2024gaussian} & &  \tbest0.50 & \tbest0.82 &  \sbest0.37 &  \tbest0.37 & 1.12 &  0.74 & 0.73 &  \sbest 1.18 & 1.29 & \sbest 0.68 & 0.77 &  \tbest0.90 & 0.42 & 0.66 &  \tbest0.49 &&  \tbest0.74 & 2h\\
 & PGSR && \best {0.36} & \best {0.57} & \tbest {0.38} & \best {0.33} & \best {0.78} & \sbest {0.58} & \best {0.50} & \best {1.08} & \best {0.63} & \best {0.59} & \best {0.46} & \best {0.54} & \best {0.30} & \best {0.38} & \best {0.34} & & \best {0.52} & \sbest 0.5h
 \\
 \bottomrule
\end{tabular}
}
\label{tab:dtu_result}
\vspace{-0.3cm}
\end{table*}

\section{Experiments}

\textbf{Datasets:} To validate the effectiveness of our method, we conducted experiments on various real-world datasets, including objects, and indoor and outdoor environments. We chose the widely used MiP-NeRF360 dataset~\cite{barron2021mip} for evaluating novel view synthesis performance. The large and complex scenes of the TnT~\cite{knapitsch2017tanks} and 15 object-centric scenes of the DTU dataset~\cite{jensen2014large} were selected to assess reconstruction quality.

\textbf{Evaluation Criterion:} We chose three widely used image evaluation metrics to validate novel view synthesis: peak signal-to-noise ratio (PSNR), structural similarity index measure (SSIM), and the learned perceptual image patch similarity (LPIPS)~\cite{zhang2018unreasonable}. For assessing surface quality, we employed the F1 score and chamfer distance.

\textbf{Implementation Details:}
Our training strategy and hyperparameters are generally consistent with 3DGS~\cite{kerbl20233d}. The training iterations for all scenes are set to 30,000. 
We adopt the densification strategy of AbsGS~\cite{ye2024absgs}. The initial value of the exposure coefficient is 0, and the learning rate is 0.001. We begin by rendering the depth for each training view, followed by utilizing the TSDF Fusion algorithm~\cite{newcombe2011kinectfusion} to generate the corresponding TSDF field. Subsequently, we extract the mesh~\cite{lorensen1998marching} from the TSDF field. We only utilize the exposure compensation on the Tanks and Temples dataset. All experiments in this paper are conducted on an Nvidia RTX 4090 GPU.

\subsection{Real-time Rendering}
For the validation of rendering quality, we follow the 3DGS method and conduct validation on the Mip-NeRF360 dataset~\cite{barron2021mip}. We compare with current state-of-the-art methods for pure novel view synthesis as well as similar reconstruction methods to ours, including NeRF~\cite{mildenhall2021nerf}, Deep Blending~\cite{hedman2018deep}, INGP~\cite{muller2022instant}, Mip-NeRF360~\cite{barron2021mip}, NeuS~\cite{wang2021neus}, 3DGS~\cite{kerbl20233d}, SuGaR~\cite{guedon2023sugar}, 2DGS~\cite{huang20242d}, and GOF~\cite{yu2024gaussian}. As shown in Table~\ref{tab:view_eval} and Fig.~\ref{fig:reconstruction2}, compared to the current state-of-the-art methods, our approach not only provides excellent surface reconstruction quality but also achieves outstanding novel view synthesis.

\begin{table}[t]\color{r}
\centering
\captionsetup{font={footnotesize}}
\caption{\textbf{Quantitative results of F1 Score$\uparrow$ for reconstruction on Tanks and Temples dataset.} PGSR achieves the best reconstruction accuracy and very fast training speed.}
\resizebox{0.98\columnwidth}{!}{
\begin{tabular}{@{}l|ccc|ccccc}
 & NeuS & Geo-Neus & Neurlangelo & SuGaR & 2D GS & GOF & PGSR\\ 
 \hline
Barn & 0.29 &  0.33 &  \best 0.70  & 0.14 & 0.45 & \tbest 0.51 &  \sbest 0.66\\
Caterpillar &  0.29 & 0.26 &  \tbest 0.36 & 0.16 & 0.24 & \sbest 0.41 & \best 0.44\\
Courthouse &  \tbest 0.17 & 0.12 &  \best 0.28 & 0.08 & 0.13 & \best 0.28 & \sbest 0.20\\
Ignatius & \sbest 0.83 & 0.72 &  \best 0.89 & 0.33 & 0.50 & 0.68 & \tbest 0.81\\
Meetingroom & 0.24 & 0.20 &  \sbest 0.32 &  0.15 & 0.18 & \sbest 0.28& \best 0.33\\
Truck &  0.45 & 0.45 &  \tbest 0.48 &  0.26 & 0.43 & \sbest 0.58& \best 0.66\\ 
\hline
Mean & 0.38 & 0.35 &  \sbest 0.50 & 0.19 & 0.32 & \tbest 0.46& \best 0.52\\
Time & \texttt{>}24h & \texttt{>}24h & \texttt{>}128h & \tbest 2h & \best 0.57h & \tbest 2h& \sbest 0.75h\\ 
\end{tabular}
}
\label{tab:tnt}
\end{table}

\subsection{Reconstruction}
We compared our method, PGSR, with current state-of-the-art neural surface reconstruction methods including NeuS~\cite{wang2021neus}, Geo-NeuS~\cite{fu2022geo}, VolSDF~\cite{Yariv:2021:Volume}, and NeuralAngelo~\cite{li2023neuralangelo}. We also compared it with recently emerged reconstruction methods based on 3DGS, such as SuGaR~\cite{guedon2023sugar}, 2DGS~\cite{huang20242d}, and GOF~\cite{yu2024gaussian}. All results are summarized in 
Fig.~\ref{fig:reconstruction2},
Fig.~\ref{fig:reconstruction_dtu},  Fig.~\ref{fig:reconstruction1}, Table~\ref{tab:dtu_result} and Table~\ref{tab:tnt}. 

\textbf{The DTU dataset:} Our method achieves the highest reconstruction accuracy with relatively fast training speed. Our method significantly outperforms other 3DGS-based reconstruction methods. As shown in Fig.~\ref{fig:reconstruction_dtu}, our surfaces are smoother and contain more details.

\textbf{The TnT dataset:} The F1 score of PGSR is similar to NeuralAngelo and better compared to other current reconstruction methods. Our training time is over 100 times faster than NeuralAngelo. Moreover, compared to NeuralAngelo, we can reconstruct more surface details. 

\begin{figure}[htb]
    \centering
    \includegraphics[width=1\linewidth]{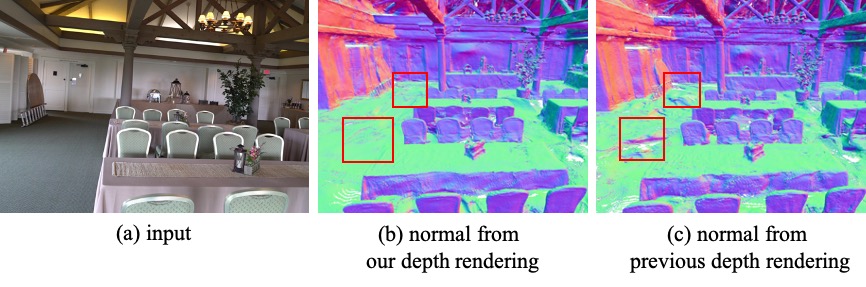}
    \captionsetup{font={footnotesize}}
    \caption{\textbf{The qualitative comparison of our unbiased depth method with the previous depth method \cite{jiang2023gaussianshader,cheng2024gaussianpro} is depicted in the normal map.} Our overall geometric structure appears smoother and more precise.}
    \label{fig:ablation_depth}
    \vspace{-0.5cm}
\end{figure}

\begin{table}\color{r}
  \captionsetup{font={footnotesize}}
  \caption{\textcolor{r}{\textbf{Ablation study on the TnT dataset.}}}
  \label{tab:ablation}
  \scriptsize%
  \centering%
  \setlength{\tabcolsep}{5pt}
  \scalebox{1.1}[1.1]{
  \begin{tabu}{c|ccc}
  \toprule
     Model setting & F1-Score$\uparrow$ & PSNR$\uparrow$ \\
  \midrule
  w/o Single-View & 0.49 & 27.02  \\
  w/o Multi-View & 0.32 & 27.30  \\
  w/o Multi-View Geometric & 0.49 & 27.07  \\
  w/o Multi-View Photometric & 0.39 & 26.83  \\
  w/o Geometric Occlusion Estimation & 0.28 & 21.70  \\
  w/o Our unbiased depth & 0.38 & 26.47  \\
  w/o Exposure Compensation & 0.49 & 25.33  \\
  \midrule
  Full model & 0.52 & 26.73 \\
  \bottomrule
  \end{tabu}
  }
  \vspace{-0.3cm}
\end{table}

\subsection{Ablations}

\textbf{Our Unbiased Depth:}
From Fig~\ref{fig:ablation_depth}, it can be observed that our overall geometric structure appears smoother and more precise, especially in flat regions. Table~\ref{tab:ablation} also demonstrates that our depth rendering method achieves higher reconstruction and rendering accuracy.


\textbf{Single-View and Multi-View Regularization:}
The single-view regularization term can provide a good initial geometric accuracy without relying on multi-view information. When single-view regularization is removed, the reconstruction accuracy decreases. Multi-view regularization constrains the consistency of geometry between multiple views, improving overall reconstruction accuracy. Both multi-view photometric and geometric consistency contribute to improving reconstruction accuracy. From Table~\ref{tab:ablation}, it is evident that multi-view regularization is crucial for reconstruction accuracy. \textcolor{r}{However, without incorporating potential occlusion estimation, the multi-view regularization term will have a negative effect, leading to poor surface reconstruction and rendering accuracy.}

The ablation results also reflect another issue: geometric constraints slightly degrade rendering quality. We speculate that this is due to an incomplete image rendering model, which forces the system to strike a balance between image and geometry losses. Further exploration may be needed to achieve synchronized improvements in geometry and novel view synthesis.


\textbf{Exposure Compensation:}
As shown in Table~\ref{tab:ablation}, exposure compensation enhances reconstruction and rendering quality.

\begin{figure}[htb]
    \centering
    \includegraphics[width=1.0\linewidth]{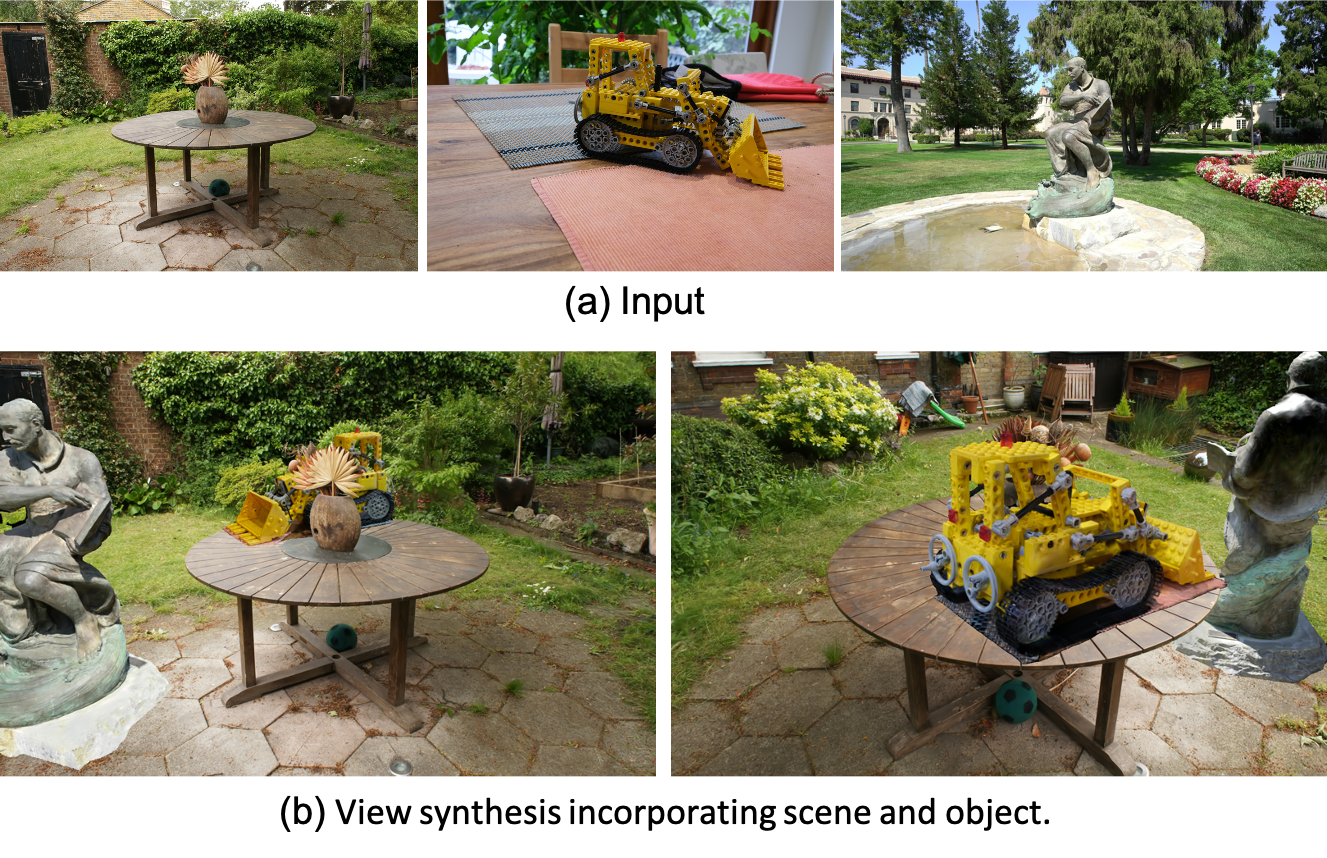}
    \captionsetup{font={footnotesize}}
    \caption{\textbf{Virtual Reality Application.} (a) Original materials, including garden scene, excavator, and Ignatius. (b) A Virtual Reality effect showcase synthesized from these original materials.}
    \label{fig:vr}
    \vspace{-0.5cm}
\end{figure}


\subsection{Virtual Reality Application}

As shown in Fig.~\ref{fig:vr}, we used our method to separately reconstruct the original materials. We then extracted the excavator and Ignatius using masks and placed them in the garden scene. By rendering the scene and objects separately and using our rendered depth to determine occlusion relationships, we achieved immersive, high-fidelity virtual reality effects with high-precision depth estimation.


\section{Limitations and Future Work}
Although our PGSR efficiently and faithfully performs geometric reconstruction, it also faces several challenges. Firstly, we cannot perform geometric reconstruction in regions with missing or limited viewpoints, leading to incomplete or less accurate geometry. Exploring methods to improve reconstruction quality under insufficient constraints using priors is another avenue for further investigation. Secondly, our method does not consider scenarios involving reflective surfaces or mirrors, so reconstruction in these environments will pose challenges. Integrating with existing 3DGS work that accounts for reflective surfaces would enhance reconstruction accuracy in such scenarios. Finally, we found that there are some floating points in the scene, which affect the rendering and reconstruction quality. Integrating more advanced 3DGS baselines~\cite{lu2023scaffold} would help further enhance the overall quality.

\section{Conclusion}
In this paper, we propose a novel unbiased depth rendering method based on 3DGS. With this method, we render the plane geometry parameters for each pixel, including normal, distance, and depth maps. We then incorporate single-view and multi-view geometric regularization, and exposure compensation model to achieve precise global consistency in geometry. We validate our rendering and reconstruction quality on the MipNeRF360, DTU, and TnT datasets. The experimental results show that our method achieves the highest geometric reconstruction accuracy and competitive rendering quality compared to state-of-the-art methods.



 
%

\bibliographystyle{plain}
\bibliography{main}


\vfill

\clearpage
\setcounter{page}{13}
\twocolumn[
    \centering
    \Large
    \textbf{PGSR: Planar-based Gaussian Splatting for Efficient and High-Fidelity Surface Reconstruction} \\
    \vspace{0.5em}Supplementary Document \\
    \vspace{1.0em}
] 
\setcounter{section}{0}
\setcounter{table}{0}
\setcounter{figure}{0}
\setcounter{equation}{0}
\renewcommand{\thetable}{\thesection\arabic{table}}
\renewcommand{\thefigure}{\thesection\arabic{figure}}
\renewcommand{\theequation}{\thesection\arabic{equation}}




\color{r}

\section{Additional Implement Details}

\subsection{Details of the Multi-View Regularization}
\textbf{Building the image graph:} We use the camera poses from the training data to calculate the relative angles and positions between all training frames, sorting the frame distances first by relative angle and then by position distance. Based on the specific characteristics of the dataset, we set the corresponding maximum number of neighboring frames, maximum relative angle, minimum relative position, and maximum relative position. For any given reference frame in the training data, other training frames that meet the set conditions are grouped into the neighboring frame set of that reference frame. For the DTU~\cite{jensen2014large}, TnT~\cite{knapitsch2017tanks}, and Mip360~\cite{barron2021mip} datasets, we uniformly set a maximum of 8 neighboring frames, a maximum relative angle of 30 degrees, a minimum relative position of 0.01, and a maximum relative position of 1.5.

\textbf{Select Neighboring Frames During Training:}
During each training iteration, we randomly select one frame from the neighboring frame set of the reference frame to be used as the neighboring frame.

\subsection{Depth Filtering}
There are often erroneous noise points near the edges of the depth map. We filter out these noise points using the angle between the rays and the normals estimated from the depth map:
\begin{equation}
    \theta=acos(\frac{\mid\bm{N}_d(\bm{p})\cdot\bm{V}(\bm{p})\mid}{\left |\bm{N}_d(\bm{p})\right | \left |\bm{V}(\bm{p})\right |}),
\end{equation}
\begin{equation}
    \bm{N}_d(\bm{p})=\frac{(\bm{P}_1-\bm{P}_0)\times(\bm{P}_3-\bm{P}_2)}{|(\bm{P}_1-\bm{P}_0)\times(\bm{P}_3-\bm{P}_2)|},
\end{equation}
For a pixel point $\bm{p}$, we sample four points from the neighboring pixels and project sampled depth points into 3D points $\{\bm{P}_j|j=1,...,4\}$ in the camera coordinate system, then calculate the normal $\bm{N}_d(\bm{p})$ for the pixel point $\bm{p}$. $\bm{V}(\bm{p})$ is the ray.

\begin{figure*}[!htb]
    \centering
    \includegraphics[width=0.7\linewidth]{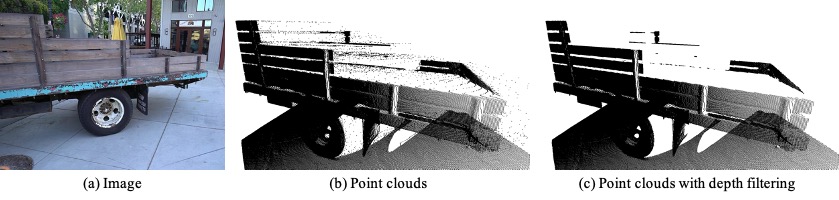}
    \captionsetup{font={footnotesize}}
    \caption{\color{r}\textbf{Qualitative comparison of point clouds without and with depth filtering.} Depth filtering can effectively remove noise at the edges.}
    \label{fig:depth_filter}
\end{figure*}

When $\theta$ is greater than $80^{\circ}$, we consider the depth point to be noise. After completing the training for each scene, we apply depth filtering as post-processing to the rendered depth maps and then perform surface reconstruction using the TSDF algorithm. Fig~\ref{fig:depth_filter} shows that noise at the edges can be filtered out. We only applied depth filtering on the TnT dataset, and according to the ablation experiment results, normal-based depth filtering slightly improves surface reconstruction accuracy.

\begin{figure*}[tb]\color{r}
    \centering
    \vspace{-0.3cm}
    \includegraphics[width=0.75\linewidth]{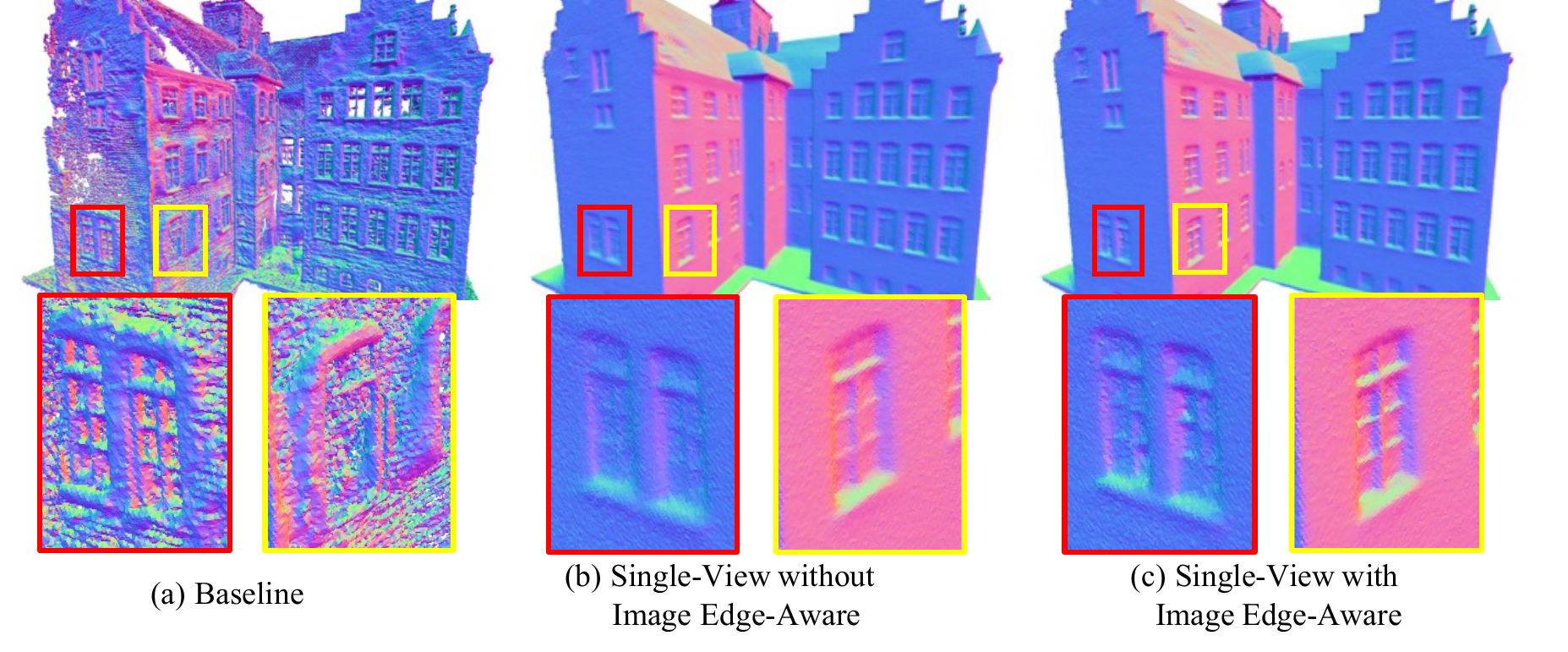}
    \captionsetup{font={footnotesize}}
    \caption{\color{r}\textbf{Qualitative comparison of surface normals under different single-view regularization terms.} The baseline model does not include any geometric regularization terms. Adding the Single-View constraint without Image Edge-Aware loss on top of the baseline can produce relatively smooth surfaces, but some areas remain insufficiently flat. Incorporating the Image Edge-Aware Single-View constraint helps retain more details.}
    \label{fig:single_view_ablation}
\end{figure*}

\begin{figure*}[tb]\color{r}
    \centering
    \includegraphics[width=0.85\linewidth]{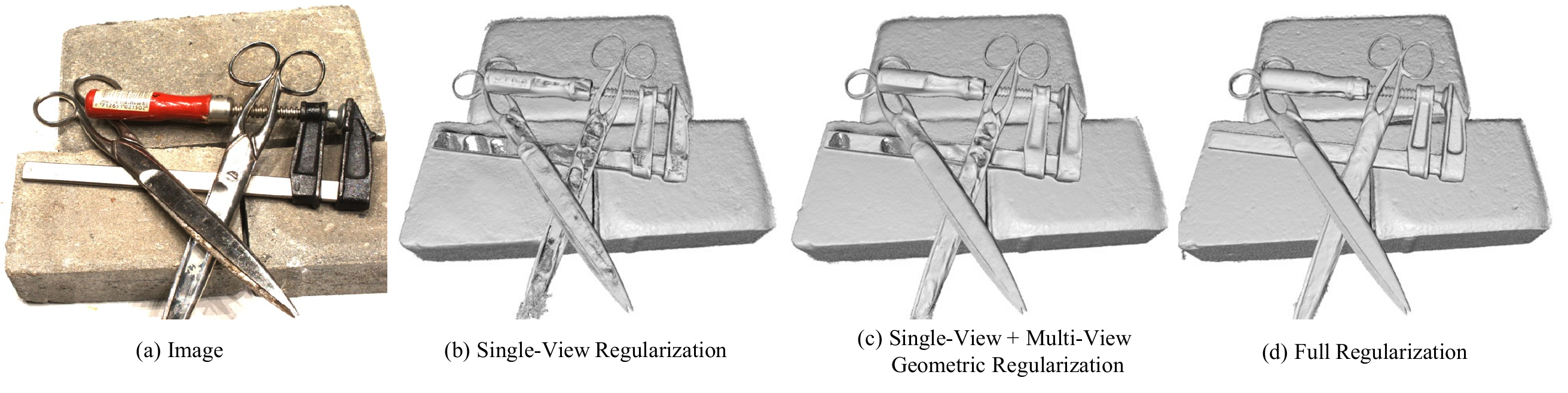}
    \captionsetup{font={footnotesize}}
    \caption{\color{r}\textbf{Qualitative comparison of different geometric regularization terms.} Although the single-view regularization term can produce relatively smooth surfaces, it tends to create holes on highly specular metallic objects. Building on this, introducing a multi-view geometric regularization term, which enforces multi-view geometric consistency, can further improve surface smoothness, but some holes may still remain. By incorporating the complete geometric regularization term, it is possible to generate surfaces that are both smooth and rich in detail.}
    \label{fig:geo_ablation}
    \vspace{-0.3cm}
\end{figure*}

\begin{figure}[tb]\color{r}
    \centering
    \includegraphics[width=1\linewidth]{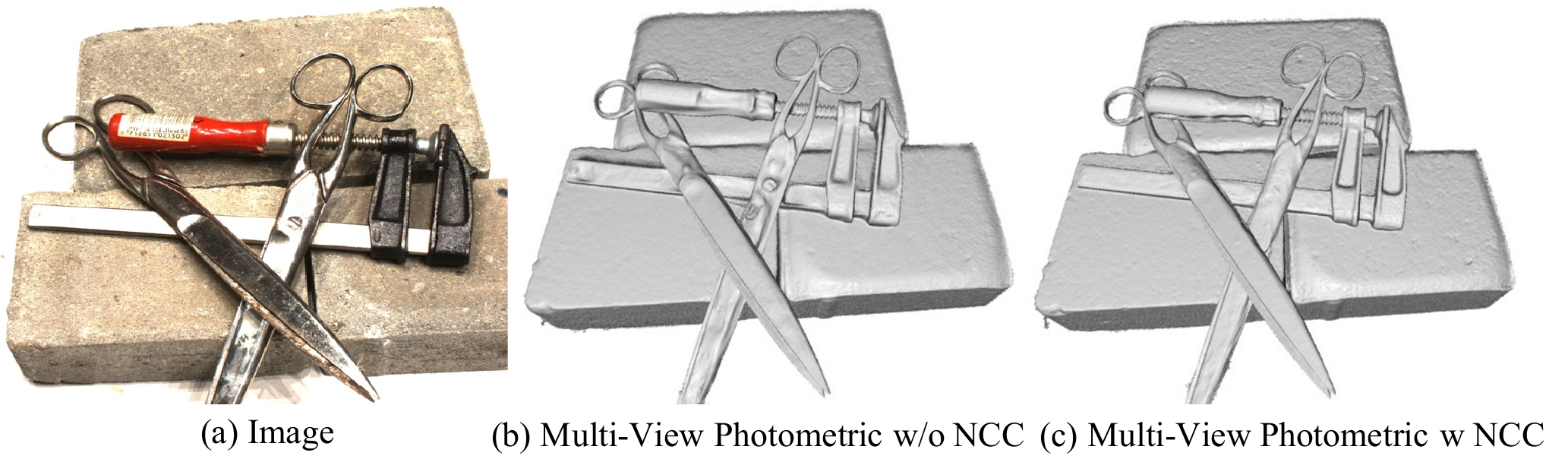}
    \captionsetup{font={footnotesize}}
    \caption{\color{r}\textbf{Qualitative comparison of multi-view photometric without and with NCC.} Multi-view photometric consistency with NCC can reconstruct higher-quality surfaces.}
    \label{fig:ncc_ablation}
\end{figure}

\begin{figure*}[tb]\color{r}
    \centering
    \includegraphics[width=0.8\linewidth]{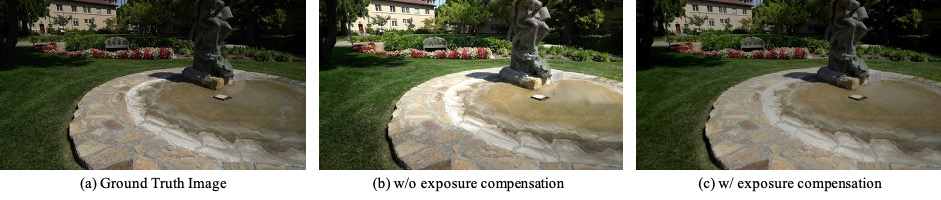}
    \captionsetup{font={footnotesize}}
    \caption{\color{r}\textbf{Qualitative comparison of rendered image without and with exposure compensation.} Exposure compensation can effectively adjust for overall brightness changes in the image. After compensation, the brightness of the adjusted image is closer to the ground truth image.}
    \label{fig:exposure_compensation}
\end{figure*}

\begin{table*}[tb]\color{r}
\centering
\captionsetup{font={footnotesize}}
\caption{\color{r}\textbf{Quantitative ablation results of F1 Score$\uparrow$ for reconstruction on Tanks and Temples dataset.}}
\centering%
\resizebox{0.75\textwidth}{!}{
\begin{tabular}{c|cccccc|c}
\toprule
   & Barn & Caterpillar & Courthouse & Ignatius & Meetingroom & Truck & Mean \\ 
 \hline

w/o Single-View & 0.64 & 0.41 & 0.18 & 0.80 & 0.30 & 0.63 & 0.49\\
w/o Edge-Aware & 0.65 & 0.44 & 0.20 & 0.81 & 0.32 & 0.65 & 0.51 \\
w/o Multi-View & 0.45 & 0.22 & 0.13 & 0.53 & 0.19 & 0.4 & 0.32 \\
w/o Multi-View Geometric & 0.65 & 0.41 & 0.16 & 0.8 & 0.32 & 0.61 & 0.49 \\
w/o Multi-View Photometric & 0.55 & 0.28 & 0.16 & 0.60 & 0.21 & 0.54 & 0.39  \\
w/o Geometric Occlusion Estimation & 0.48 & 0.03 & 0.17 & 0.44 & 0.18 & 0.40 & 0.28  \\
w/o Our Unbiased Depth & 0.5 & 0.31 & 0.18 & 0.70 & 0.15 & 0.45 & 0.38\\
w/o Exposure Compensation & 0.66 & 0.42 & 0.20 & 0.77 & 0.29 & 0.62 & 0.49 \\
w/o Depth Filter& 0.66 & 0.43 & 0.20 & 0.80 & 0.33 & 0.60 & 0.50 \\
\midrule
Full model & 0.66 & 0.44 & 0.20 & 0.81 & 0.33 & 0.66 & 0.52 \\
\bottomrule

\end{tabular}
}
\label{tab:detailed_tnt_f1}
\end{table*}
\begin{table*}[tb]\color{r}
\centering
\captionsetup{font={footnotesize}}
\caption{\color{r}\textbf{Quantitative ablation results of PSNR$\uparrow$ for reconstruction on Tanks and Temples dataset.}}
\centering%
\resizebox{0.75\textwidth}{!}{
\begin{tabular}{c|cccccc|c}
\toprule
   & Barn & Caterpillar & Courthouse & Ignatius & Meetingroom & Truck & Mean \\ 
 \hline

w/o Single-View & 29.45 & 27.06 & 23.89 & 26.68 & 28.14 & 26.9 & 27.02\\
w/o Edge-Aware & 29.14 & 26.80 & 23.43 & 26.49 & 27.67 & 26.58 & 26.69 \\
w/o Multi-View & 29.82 & 27.11 & 24.3 & 26.89 & 28.51 & 27.14 & 27.30 \\
w/o Multi-View Geometric & 29.55 & 27.05 & 24.14 & 26.63 & 28.16 & 26.90 & 27.07 \\
w/o Multi-View Photometric & 29.33 & 26.91 & 23.62 & 26.61 & 27.75 & 26.76 & 26.83  \\
w/o Geometric Occlusion Estimation & 22.44 & 21.20 & 19.95 & 21.95 & 22.52 & 22.51 & 21.70  \\
w/o Our Unbiased Depth & 28.60 & 26.63 & 22.88 & 26.49 & 27.63 & 26.58 & 26.47\\
w/o Exposure Compensation & 29.21 & 24.27 & 22.47 & 23.17 & 26.19 & 26.67 & 25.33 \\
\midrule
Full model & 29.2 & 26.81 & 23.48 & 26.48 & 27.72 & 26.66 & 26.73 \\
\bottomrule

\end{tabular}
}
\label{tab:detailed_tnt_psnr}
\end{table*}
\begin{table*}[tb]\color{r}
\centering
\captionsetup{font={footnotesize}}
\caption{\color{r}\textbf{Quantitative ablation results of F1 Score$\uparrow$ for reconstruction on Tanks and Temples dataset.}}
\centering%
\resizebox{0.75\textwidth}{!}{
\begin{tabular}{c|cccccc|c}
\toprule
   & Barn & Caterpillar & Courthouse & Ignatius & Meetingroom & Truck & Mean \\ 
\hline

2DGS~\cite{huang20242d} & 0.45 & 0.24 & 0.13 & 0.50 & 0.18 & 0.43 & 0.32 \\
2DGS~\cite{huang20242d} + Multi-View & 0.51 & 0.33 & 0.10 & 0.67 & 0.24 & 0.61 & 0.41 \\
2DGS with expected depth~\cite{huang20242d} + Multi-View & 0.64 & 0.39 & 0.13 & 0.75 & 0.31 & 0.61 & 0.47 \\
\bottomrule

\end{tabular}
}
\label{tab:ablation_2dgs}
\end{table*}

\section{Additional Results}

\subsection{Additional Detailed Ablation Experiments}

Table~\ref{tab:detailed_tnt_f1} shows the ablation results for each scene in the TnT dataset. It can be seen that each modification is necessary and contributes to improving surface reconstruction accuracy.

\textbf{Image Edge-Aware Single-View:}
From Table~\ref{tab:detailed_tnt_f1} and \ref{tab:detailed_tnt_psnr}, it can be observed that single-view constraints help improve surface reconstruction accuracy but slightly reduce rendering quality. Edge awareness offers only a slight improvement in surface reconstruction, but as seen in the comparison between (c) and (b) in Fig~\ref{fig:single_view_ablation}, edge awareness is beneficial for preserving more details.

\textbf{Multi-View Geometric Consistency:}
The qualitative effect of multi-view geometric consistency is shown in Fig~\ref{fig:geo_ablation}. By enforcing multi-view geometric consistency, surface smoothness is improved. For highly specular metallic objects, it is difficult to ensure geometric consistency relying on images and single-view constraints. By imposing multi-view geometric consistency, the reconstructed surface becomes more complete and smooth. Table~\ref{tab:detailed_tnt_f1} also shows that multi-view geometric consistency improves surface reconstruction accuracy across various scenes, though it slightly reduces rendering quality, as demonstrated in Table~\ref{tab:detailed_tnt_psnr}.

\textbf{Multi-View Photometric Consistency:}
The qualitative results of multi-view photometric consistency are shown in Figs~\ref{fig:geo_ablation} and~\ref{fig:ncc_ablation}. Fig~\ref{fig:geo_ablation} demonstrates that the proposed multi-view photometric consistency significantly improves surface reconstruction quality, producing smooth and detail-rich surfaces, especially in highly specular scenes. Fig~\ref{fig:ncc_ablation} illustrates the benefits of using NCC (Normalized Cross-Correlation) to compute photometric consistency. In highly specular scenes, directly measuring photometric consistency with the absolute difference between two pixel patches is sensitive to pixel intensity variations, resulting in poor surface reconstruction quality. In contrast, NCC calculates similarity based on correlation coefficients, making it less sensitive to brightness changes and more suitable for image matching, thereby achieving better surface quality. Table~\ref{tab:detailed_tnt_f1} and \ref{tab:detailed_tnt_psnr} shows that in various scenes, the multi-view photometric consistency constraint plays a crucial role in improving reconstruction quality, though it slightly reduces rendering quality.

\textbf{Unbiased Depth:}
Unbiased depth rendering is crucial for producing high-quality surface reconstructions. As shown in Table~\ref{tab:detailed_tnt_f1}, unbiased depth rendering is essential for improving surface reconstruction quality across all scenes. Table~\ref{tab:detailed_tnt_psnr} also demonstrates that unbiased depth rendering can slightly enhance rendering quality.

\textbf{Exposure Compensation:}
We assign two exposure coefficients to each frame to model the exposure variation across frames. The qualitative results of the Ignatius sequence from the Tanks and Temples dataset are shown in Fig~\ref{fig:exposure_compensation}. Most frames of the Ignatius sequence are relatively bright, with a few being darker. Without exposure compensation, the rendered images tend to appear overly bright, whereas the images with exposure compensation have overall brightness closer to the ground truth. As seen from Table~\ref{tab:detailed_tnt_f1} and Table~\ref{tab:detailed_tnt_psnr}, exposure compensation helps improve both surface reconstruction and rendering quality.

\textbf{Depth Filter:}
For the Tanks and Temples dataset, after training each scene, depth maps for each frame are rendered. Depth filtering is then applied to remove depth noise, followed by the use of the TSDF algorithm to fuse the cleaned depth maps and generate an SDF field, from which the mesh is extracted. Table~\ref{tab:detailed_tnt_f1} shows that depth filtering can slightly improve surface reconstruction quality. The ablation study Fig~\ref{fig:depth_filter} demonstrates the effectiveness of depth filtering.

\textbf{Related to 2DGS:}
For TnT scene, 2DGS~\cite{huang20242d} defaults to using median depth. 2DGS with expected depth, on the other hand, employs the expected depth mode. We conducted ablation comparisons between 2DGS and our multi-view regularization. As shown in the Table~\ref{tab:ablation_2dgs}, our multi-view regularization applies to 2DGS, significantly enhancing the surface reconstruction accuracy of 2DGS. However, due to the 'disk-aliasing' problem associated with median depth, the accuracy improvement is relatively limited compared to expected depth. Even with the addition of multi-view constraints, the surface accuracy of 2DGS is still inferior to that of PGSR.

\color{black}
\subsection{More Results}
\begin{figure*}[t]
    \centering
    \includegraphics[width=0.9\linewidth]{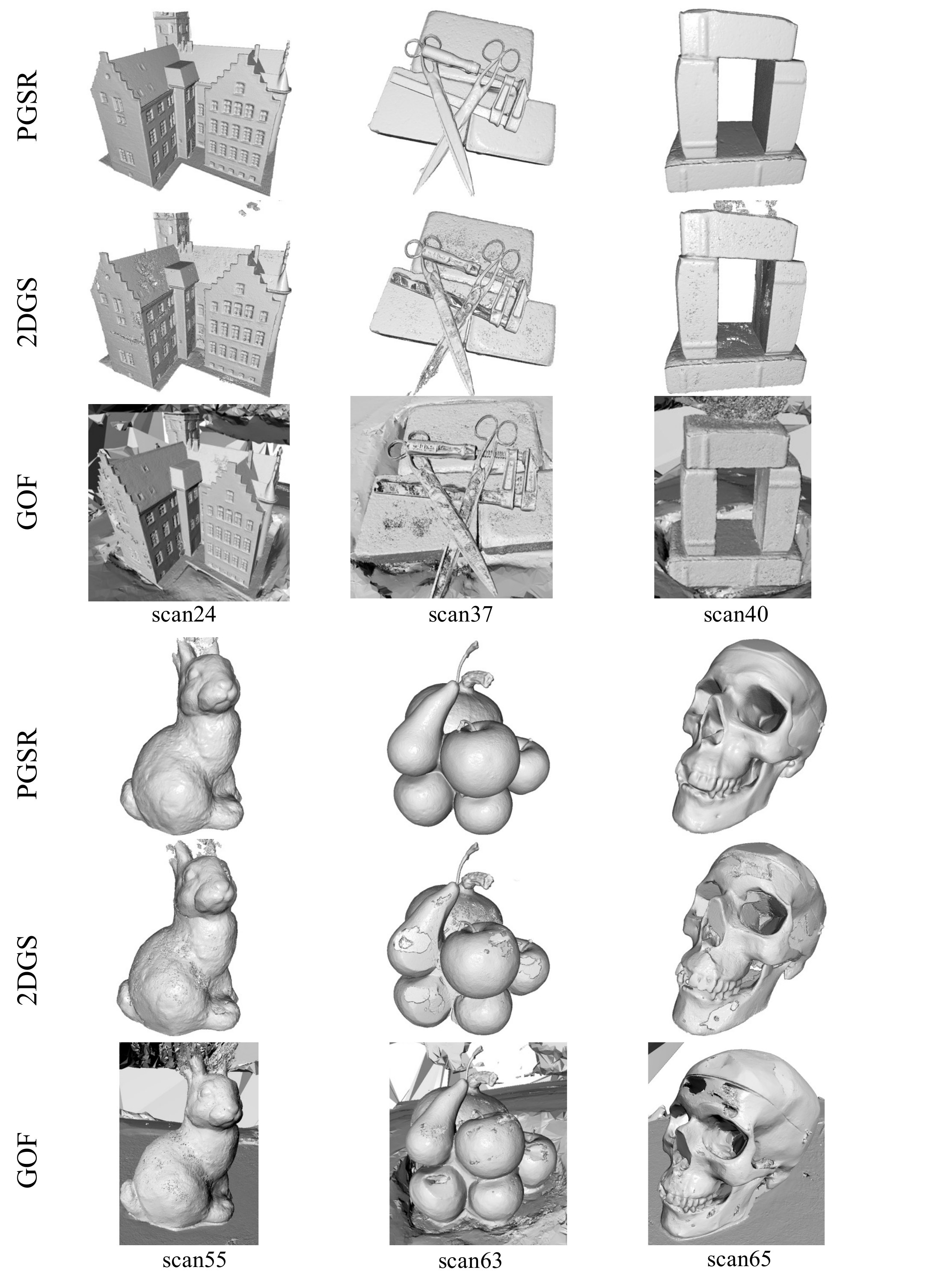}
    \captionsetup{font={footnotesize}}
    \caption{
    Qualitative comparisons in surface reconstruction between PGSR, 2DGS~\cite{huang20242d}, and GOF~\cite{yu2024gaussian} on the DTU dataset.
    }
    \label{fig:dtu3}
\end{figure*}

\begin{figure*}[t]
    \centering
    \includegraphics[width=0.9\linewidth]{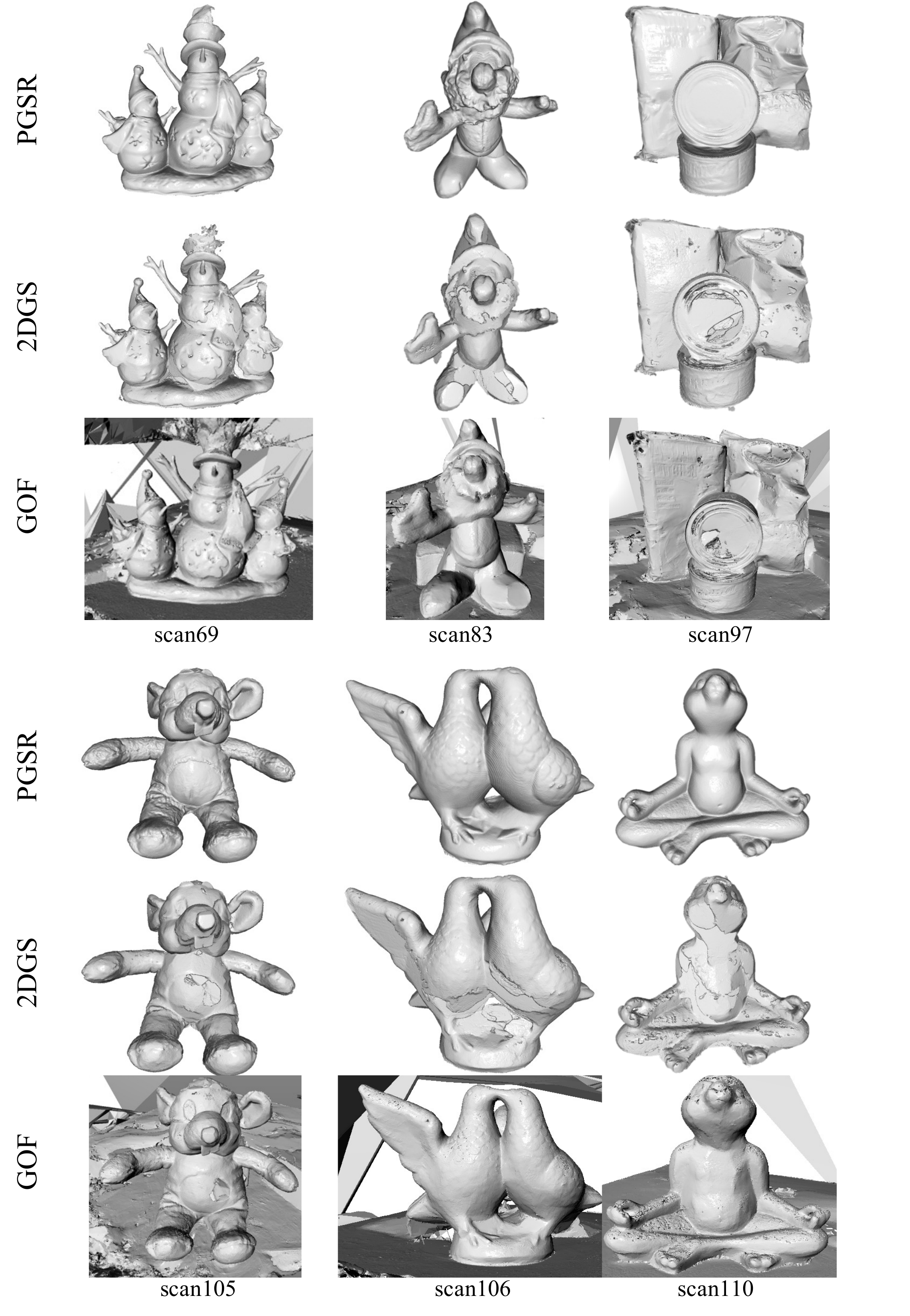}
    \captionsetup{font={footnotesize}}
    \caption{
    Qualitative comparisons in surface reconstruction between PGSR, 2DGS, and GOF on the DTU dataset.
    }
    \label{fig:dtu4}
\end{figure*}

\begin{figure*}[t]
    \centering
    \includegraphics[width=0.9\linewidth]{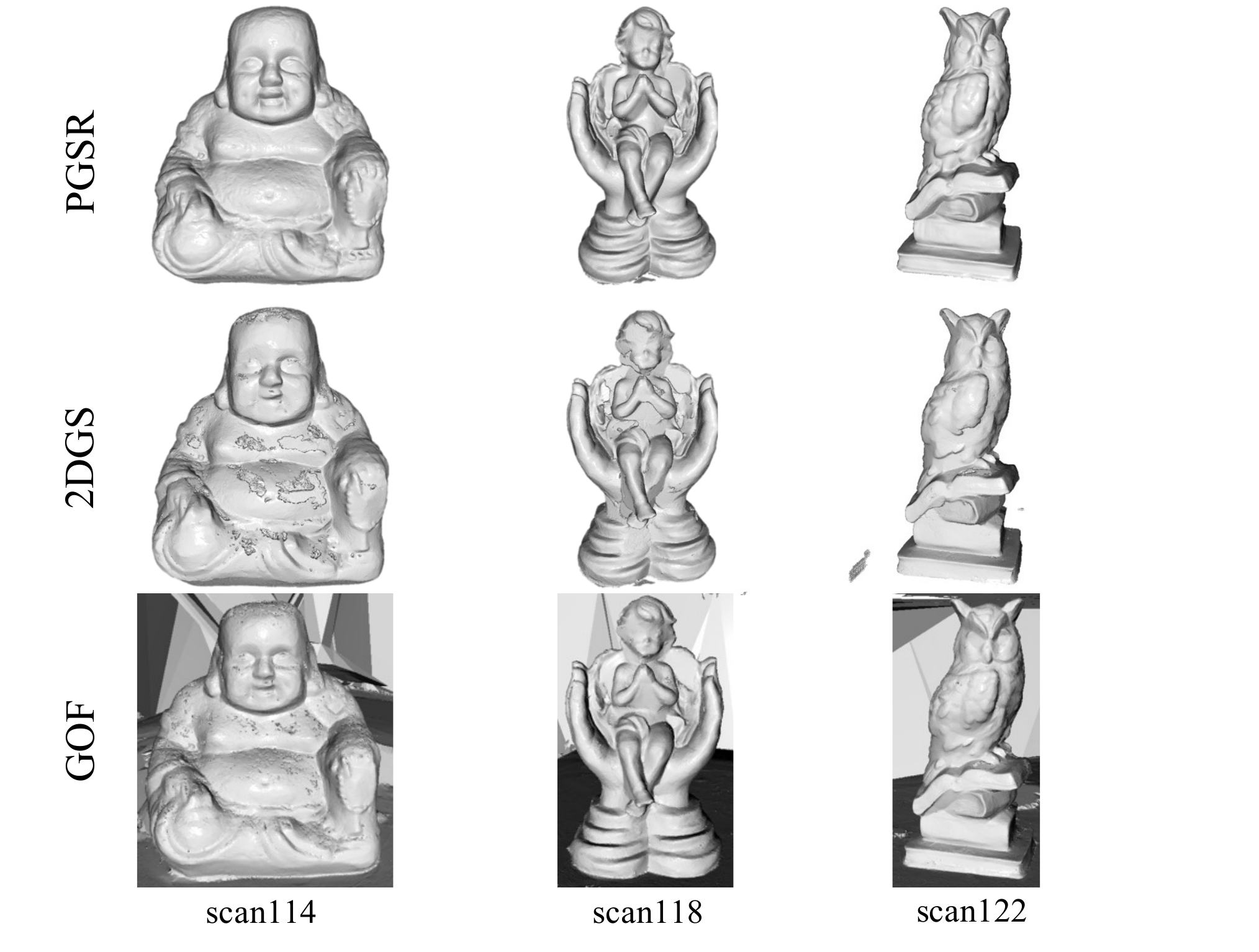}
    \captionsetup{font={footnotesize}}
    \caption{
    Qualitative comparisons in surface reconstruction between PGSR, 2DGS, and GOF on the DTU dataset.
    }
    \label{fig:dtu5}
\end{figure*}

\begin{figure*}[t]
    \centering
    \includegraphics[width=0.9\linewidth]{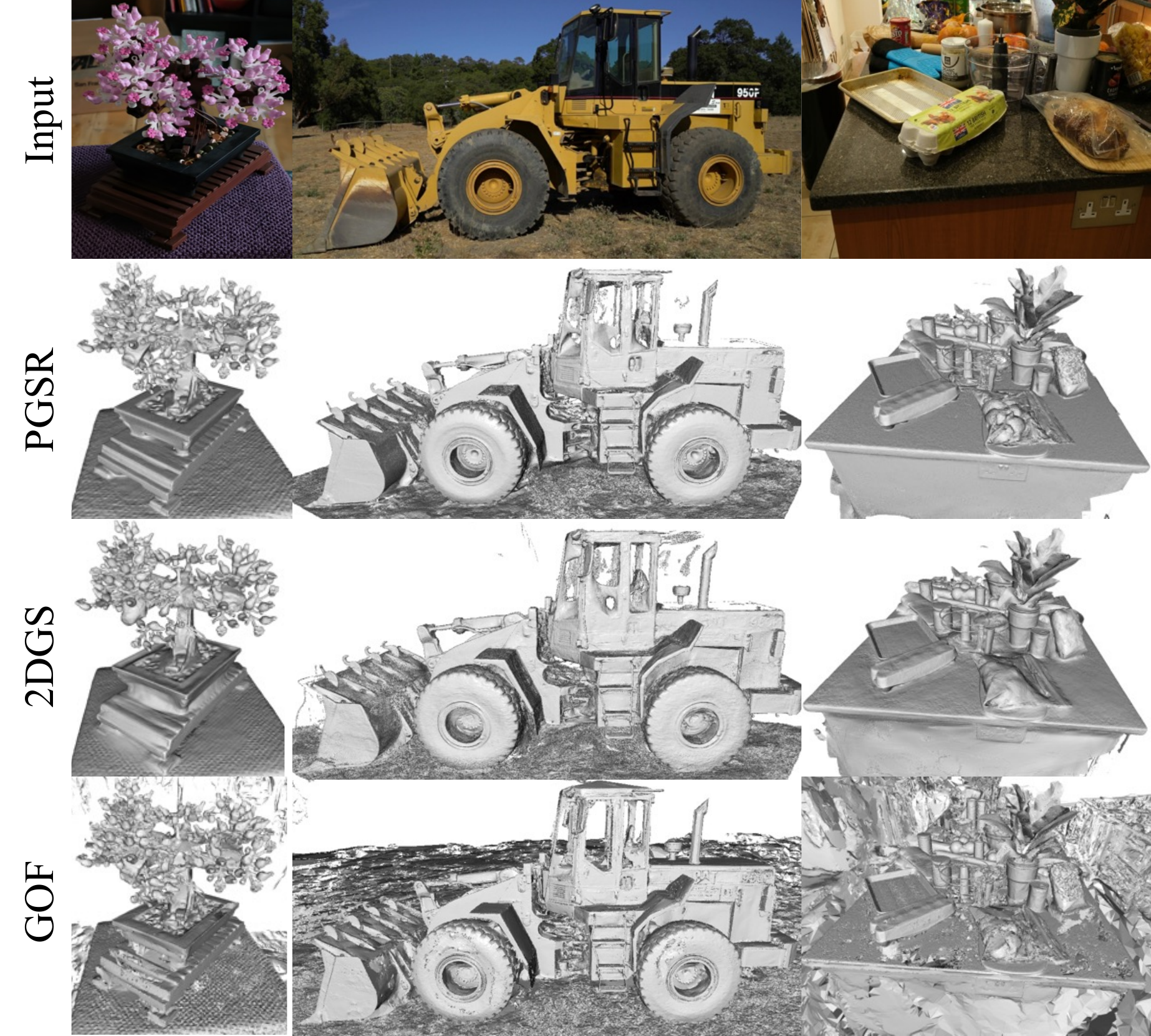}
    \captionsetup{font={footnotesize}}
    \caption{
    Qualitative comparisons in surface reconstruction between PGSR, 2DGS, and GOF.
    }
    \label{fig:dtu6}
\end{figure*}

\begin{figure*}[t]
    \centering
    \includegraphics[width=0.9\linewidth]{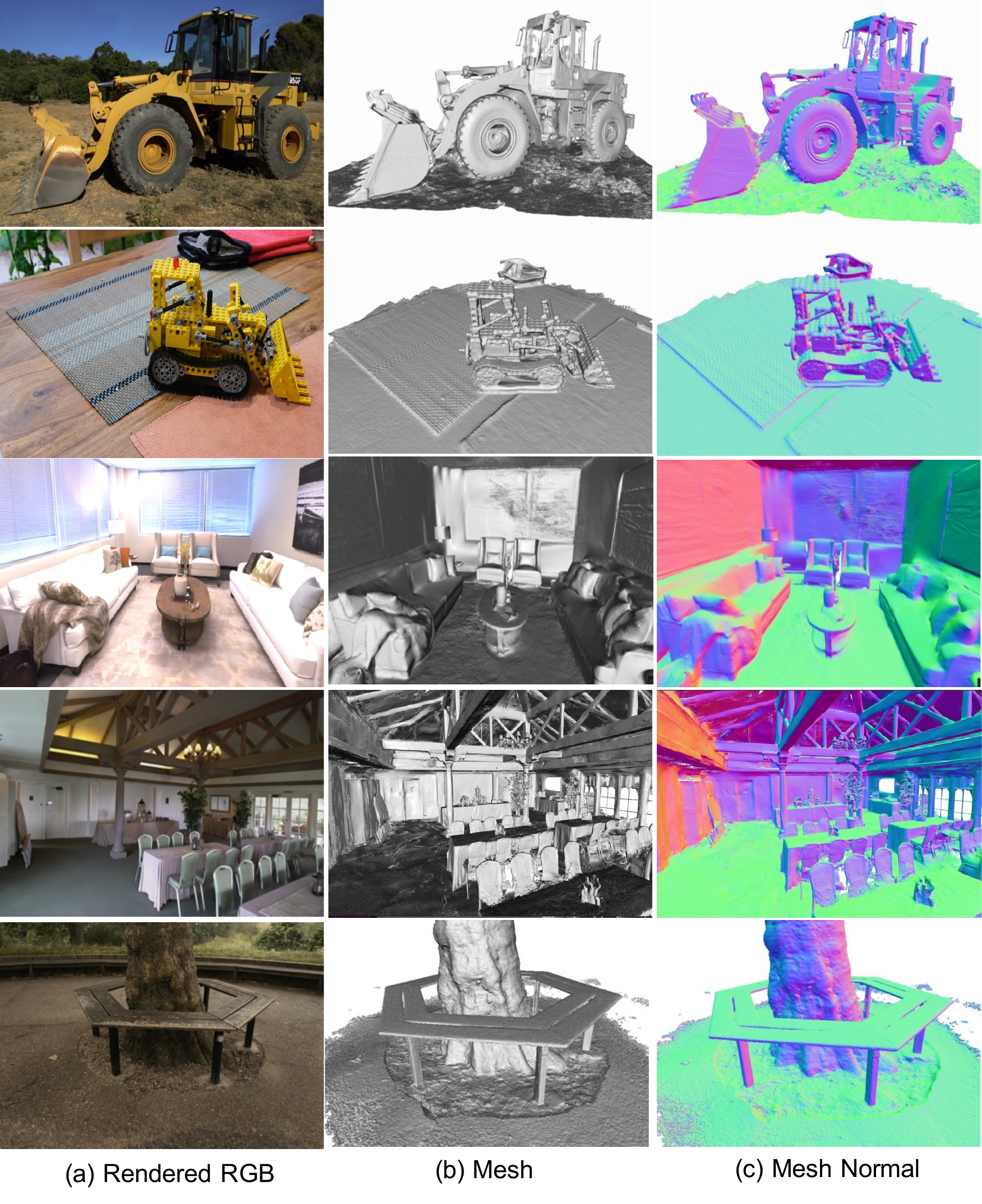}
    \captionsetup{font={footnotesize}}
    \caption{
    PGSR achieves high-precision geometric reconstruction in various indoor and outdoor scenes from a series of RGB images without requiring any prior knowledge.
    }
    \label{fig:reconstruction3}
\end{figure*}

Figs~\ref{fig:dtu3} to \ref{fig:dtu6} present a comparison of surface reconstruction results across various scenes between PGSR, 2DGS~\cite{huang20242d}, and GOF~\cite{yu2024gaussian}. We provide additional results on various scenes and objects, further confirming the capability of our method, PGSR, in achieving high-fidelity geometric reconstruction, as shown in Fig~\ref{fig:reconstruction3}.

\end{document}